\documentclass[journal]{IEEEtran} %\documentclass[10pt,draftcls,onecolumn]{IEEEtran}

\usepackage{amssymb} 
\usepackage{amsmath}
\usepackage{etoolbox}
\usepackage{amssymb}
\usepackage{graphicx}
\newtheorem{definition}{\textbf{Definition}}

\newtheorem{proposition}{\textbf{Proposition}}
\newtheorem{algorithm}{\textbf{Algorithm}}

\newtheorem{corollary}{\textbf{Corollary}}
\newtheorem{lemma}{\textbf{Lemma}}

\usepackage{algorithm}
\newcommand{\mycomment}[1]{}

\usepackage{algorithmic}

\newtheorem{remark}{\textbf{Remark}}
\newtheorem{example}{\textbf{Example}}
\usepackage[none]{hyphenat}  
\newcommand{\bld}[1]{\boldsymbol{#1}}

\author{Ioannis  Kordonis,~~ Petros Maragos\thanks{ The authors are with the National Technical University of Athens, School of Electrical and Computer Engineering
9, Iroon Polytechniou St
Athens, Postal Code 157 80, Greece.\newline Author emails: I. Kordonis: kordonis@central.ntua.gr,  P. Maragos maragos@cs.ntua.gr.}
\thanks{ The research project was supported by the Hellenic Foundation for Research and Innovation (H.F.R.I.) under the ``2nd Call for H.F.R.I. Research Projects to support Faculty Members \&  Researchers''  (Project Number:2656, Acronym: TROGEMAL).}}

\title{Revisiting Tropical Polynomial Division: Theory, Algorithms and Application to Neural Networks.}

\begin{document}
\maketitle

\begin{abstract} 
Tropical geometry has recently found several applications in the analysis of neural networks with piecewise linear activation functions. This paper presents a new look at the problem of tropical polynomial division and its application to the simplification of neural networks. We analyze tropical polynomials with real coefficients, extending earlier ideas and methods developed for polynomials with integer coefficients. We first prove the existence of a unique quotient-remainder pair and characterize the quotient in terms of the convex bi-conjugate of a related function. Interestingly, the quotient of tropical polynomials with integer coefficients does not necessarily have integer coefficients. Furthermore, we develop a relationship of tropical polynomial division with the computation of the convex hull of unions of convex polyhedra and use it to derive an exact algorithm for tropical polynomial division. An approximate algorithm is also presented, based on an alternation between data partition and linear programming. We also develop special techniques to divide composite polynomials, described as sums or maxima of simpler ones. Finally, we present some numerical results to illustrate the efficiency of the algorithms proposed, using the MNIST handwritten digit and CIFAR-10 datasets. \sloppy
\sloppy

\end{abstract}

\section{Introduction}

Tropical geometry is a relatively new research field combining elements and ideas from polyhedral and algebraic geometry \cite{maclagan2009introduction}. The underlying algebraic structure is the max-plus semiring (also known as tropical semiring), where the usual addition is replaced by maximization and the standard multiplication by addition. Tropical polynomials (also known as max-polynomials) are the polynomials in the max-plus algebra and have a central role in tropical geometry. This work deals with the division of tropical polynomials and presents some applications in neural network compression. \sloppy

A link between  tropical geometry and machine learning was recently developed \cite{maragos2021tropical,zhang2018tropical}. An important application is the analysis  of neural networks with piecewise linear activations (e.g., ReLU). In this front, \cite{zhang2018tropical,charisopoulos2018tropical,montufar2021sharp, alfarra2022decision,hertrich2021towards} use a tropical representation of neural networks to describe the complexity of a network structure, defined as the number of its linear regions, or describe their decision boundaries. \cite{trimmel2020tropex} presents an algorithm to extract the linear regions of deep ReLU neural networks.   Papers \cite{smyrnis2019tropical,smyrnis2020maxpolynomial, smyrnis2020multiclass}  deal with the problem of tropical polynomial division and its use in the simplification of neural networks. The analysis is, however, restricted to polynomials with integer coefficients. Article \cite{misiakos2021neural} uses  tropical representations of neural networks and employs polytope approximation tools to simplify them. Another class of networks represented in terms of tropical algebra is morphological neural networks \cite{ritter1996introduction,ritter2003lattice,sussner2011morphological,charisopoulos2017morphological,shen2019deep,dimitriadis2021advances}. Other applications of tropical algebra and geometry include the tropical modeling of classical algorithms in probabilistic graphical models  \cite{8445777}, \cite{8683127} and piecewise linear regression \cite{maragos2019tropical, maragos2020multivariate}. Neural networks employing the closely related log-sum-exp nonlinearity were presented in \cite{calafiore2019log,calafiore2020universal}.

Another problem very much related to the current work is the factorization of tropical polynomials. This problem  was first studied in \cite{kim2005factorization,grigg2007elementary,gao2001decomposition,tiwary2008hardness}, for the single-variable case, and in \cite{lin2017linear, crowell2019tropical} for the multivariate case. The problem of tropical rational function simplification was studied in \cite{tran2022minimal}. Another related problem is approximation in tropical algebra (see, e.g., \cite{akian2011best}). \sloppy

\textit{Contribution:} This work deals with the problem of tropical polynomial division and its applications in simplifying neural networks.
The analysis generalizes previous works \cite{smyrnis2019tropical,smyrnis2020maxpolynomial, smyrnis2020multiclass} to include polynomials with real coefficients. We first introduce a new notion of division and show that there is a unique quotient-remainder pair. Furthermore, we show that the quotient is equal to the convex bi-conjugate of the difference between the dividend and the divisor. Then, the quotient is characterized in terms of the closed convex hull of the union of a finite collection of convex (unbounded) polyhedra. This characterization leads to a simple exact algorithm based on polyhedral geometry. We propose an efficient approximate scheme inspired by an optimization algorithm for convex, piecewise-linear fitting \cite{magnani2009convex}. The approximate algorithm alternates between data clustering and linear programming. Then, we focus on developing algorithms for dividing composite polynomials expressed as the sum of simpler ones. The division results are then applied to simplify neural networks with ReLU activation functions. The resulting neural network has fewer neurons with maxout activations and a reduced total number of parameters. Finally, we present numerical results for the MNIST handwritten digit recognition and the CIFAR-10 image classification problem. As a baseline for comparison, we use the structured L1 pruning without retraining. The proposed method is very competitive in the large compression regime, that is, in the case where there are very few parameters remaining after compression.

The rest of the paper is organized as follows: Section \ref{Sec:Preliminaries} presents some preliminary algebraic and geometric material. In Section  \ref{Sec:Theory}, we develop some theoretical tools for tropical polynomial division. An exact division algorithm is presented in  Section \ref{Sec:ExactAlg}, and an approximate algorithm  in Section
\ref{Sec:ApproxAlg}. Section \ref{Sec:Composite} deals with the division of composite tropical polynomials. Finally,
\ref{Sec:Numerical} gives some numerical examples. The proofs of the theoretical results are presented in the Appendix.

\section{Preliminaries, Background and notation} 
\label{Sec:Preliminaries}
Max-plus algebra is a modification of the usual algebra of the real numbers, where the usual summation is substituted by maximum and the usual multiplication is
substituted by summation \cite{baccelli1992synchronization,heidergott2014max,butkovivc2010max,cuninghame2012minimax}. Particularly, max-plus algebra is defined on the set $\mathbb{R}_{\max}=\mathbb{R}\cup\{-\infty\}$ with the binary operations ``$\vee$'' and ``$+$". The operation ``$+$" is the usual addition. The operation ``$\vee$" stands for the maximum, i.e. for $x,y\in\mathbb{ R}_{\max}$, it holds $x\vee y = \text{max} \{x,y\}$. For more than two terms we use``$\bigvee$'', i.e.,  $\bigvee_{i\in I} x_i  = \sup \{x_i:i\in I\}$. We will also use the `$\wedge $' notation to denote the minimum, i.e.,  $x\wedge y = \text{min} \{x,y\}$. 
A unified view of max-plus and several related algebras, was developed in \cite{maragos2017dynamical}, in terms of  weighted lattices.

A tropical polynomial is a polynomial in the max-plus algebra. Particularly, a tropical polynomial function \cite{maclagan2021introduction,joswig2021essentials,itenberg2009tropical} is defined as $p:\mathbb{R}^n\rightarrow \mathbb{R}_{\text{max}}$ with
\begin{equation}
p(\boldsymbol x) = \bigvee_{i=1}^{m_p} (\boldsymbol{a}_i^T\boldsymbol{x}+b_i),\label{Pdef}
\end{equation}
where $ \boldsymbol a_i\in \mathbb R^n$ and $b_i\in \mathbb R_{\text{max}}$. In this paper, we will use the terms tropical polynomial and tropical polynomial function interchangeably.

\begin{figure}
\centering
\includegraphics[width=0.25\textwidth]{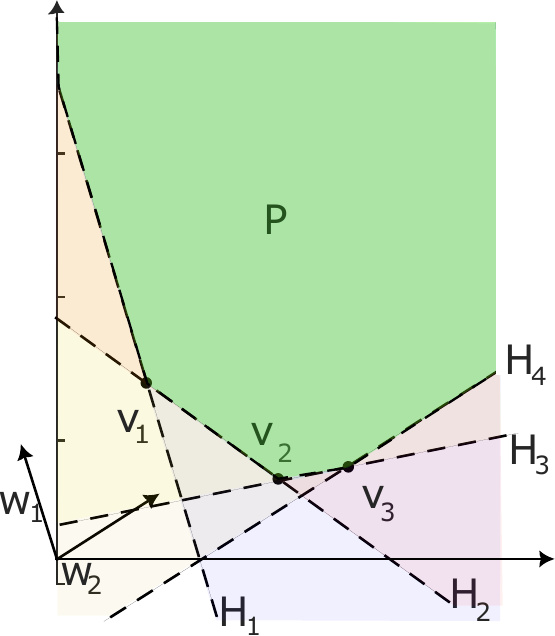}
\caption{\textit{Illustration of the Minkovski-Weil theorem. The unbounded polyhedron P (green color in the figure) has two representations: $\mathcal H-$reprsentation as the intersection of halfspaces $H_1,\dots,H_4$ and $\mathcal V-$representation generated by vertices $v_1,v_2,v_3$ and rays $w_1,w_2$.}}
\label{koookkk}
\end{figure}

For the tropical polynomial $p(\bld x),$ the Newton polytope is defined as 
\begin{equation}
\text{Newt}(p) =\text{ conv}\{\boldsymbol{a}_1,\dots,\boldsymbol{a}_{m_p}\},
\end{equation}
where ``conv'' stands for the convex hull,  
and extended Newton polytope as
\begin{equation}
\text{ENewt}(p) =\text{ conv}\{[\bld a_1^T~b_1]^T,\dots,[\boldsymbol a_{m_p}^T b_{m_p}]^T\}.
\end{equation}
The values of the polynomial function $p$ depend on the upper hull of the extended Newton polytope \cite{maclagan2009introduction} (see also \cite{charisopoulos2018tropical}).  
A polyhedron is subset of $\mathbb R^n$ which can be represented either as the set of solutions of a finite collection of linear inequalities ($\mathcal H -$representation)
$$P =\{\bld x\in\mathbb R^n: \bld A\bld x\leq \bld b\},$$
or  in terms of a set of vertices $\bld v_1,\dots,\bld v_l\in \mathbb R^n$ and a set of rays $\bld w_1,\dots,\bld w_s\in \mathbb R^n$ as
\begin{align}
P = \left\{\sum_{j=1}^{l} \lambda_j \bld v_j+\sum_{j=1}^{s} \mu_j \bld w_j:\lambda_j\geq0,\sum_{j=1}^{s} \lambda_j=1,  \mu_j\geq0\right\},
\label{resol_equ}
\end{align}
($\mathcal V-$representation). The  Minkowski-Weyl (resolution) theorem states that each polyhedron admits both representations \cite{fukuda2016lecture}. Figure \ref{koookkk} illustrates the  Minkowski-Weyl resolution of an unbounded polyhedron. the  Furthermore, there are several well-known algorithms for representation conversion (e.g., \cite{fukuda2016lecture}, Ch. 9). Bounded polyhedra are called polytopes. It is often convenient to describe polytopes in terms of extended representations (e.g., \cite{kaibel2011extended}). An extension of a polytope $P\subset\mathbb R^n$ is a polyhedron $Q\subset\mathbb R^{n'}$, with $n'>n$, along with a linear projection $F:R^{n'}\rightarrow \mathbb R^n$ such that $F(Q)=P$.

For a function $f:\mathbb R^n \rightarrow\mathbb R\cup\{\infty\}$ the epigraph is defined as
$$\text{epi}(f)=\{(\bld x,z)\in\mathbb R^{n+1}: z\geq f(\bld x)\}.$$
The convex conjugate of $f$ is a function $f^\star:\mathbb R^n \rightarrow\mathbb R\cup\{+\infty,-\infty\}$ given by (see \cite{rockafellar2015convex})
$$f^\star(\bld a)=\sup\{\bld a^T \bld x -f(\bld x):\bld x\in \mathbb R^n\}.$$

%Similarly, the hypograph is defined as$$\text{hypo}(f)=\{(\bld x,z)\in\mathbb R^{n+1}: z\leq f(\bld x)\}.$$

For a pair of polyhedra $P_1, P_2\subset\mathbb R^n$ the  Minkowski sum is defined as $$P_1  \oplus P_2 = \{x+y:x\in P_1,y\in P_2\}.$$

The following properties hold (see \cite{charisopoulos2018tropical})
\begin{align*}
\text{Newt}(p_1+p_2) &=\text{Newt}(p_1)\oplus\text{Newt}(p_2), \\
\text{Newt}(p_1\vee p_2) &=\text{conv}(\text{Newt}(p_1)\cup\text{Newt}(p_2) ).
\end{align*}
Similar relations hold for the extended Newton polytopes as well.

 We then introduce a dual function of a tropical polynomial, called  Extended Newton Function (ENF).  For a tropical polynomial $p$, the ENF  is defined as:
$$\text{ENF}_p(\bld a) = \sup\{b\in \mathbb R: [\bld a^T~b]^T\in \text{ENewt}(p) \},$$
where $\bld a\in \mathbb R^n$ and $\sup(\emptyset)=-\infty$. This function describes the upper hull of the extended Newton polytope. Thus, the ENF fully characterizes the tropical polynomial function $p$.

\begin{proposition}
\label{LemmaBackground}
Let $p_1,p_2$ be two tropical polynomials. Then
 $p_1(\bld x)\leq p_2(\bld x) $, for all $\bld x\in \mathbb R^n$, if and only if $\text{ENF}_{p_1}(\bld a)\leq \text{ENF}_{p_2}(\bld a)$, for all $\bld a\in \mathbb R^n$. 
\end{proposition}
\textit{Proof}: See Appendix \ref{monotonic_Prop}.
   
The significance of ENFs is that we can often express inequality $\text{ENF}_{p_1}\leq \text{ENF}_{p_2}$, using a finite number of inequalities in terms of the tropical polynomial coefficients of $p_1$, $p_2$.

\mycomment{
\begin{lemma}
\label{LemmaBackground}
Let $p_1,p_2$ be two tropical polynomials. Then:
\begin{itemize}
\item[(i)] It holds $p_1(\bld x)\leq p_2(\bld x) $, for all $\bld x\in \mathbb R^n$, if and only if $\text{ENF}_{p_1}(\bld a)\leq \text{ENF}_{p_2}(\bld a)$, for all $\bld a\in \mathbb R^n$. \item[(ii)] The function $\text{ENF}_{p_j}(\bld a)$ is piecewise linear and concave. 
\item[(iii)] It holds
\begin{align*}
\text{hypo} (\text{ENF}_{p_1+ p_2}) &=\text{hypo} (\text{ENF}_{p_1}) \oplus\text{hypo} (\text{ENF}_{p_2})\\
\text{hypo} (\text{ENF}_{p_1\vee p_2}) &=\text{conv}(\text{hypo} (\text{ENF}_{p_1}) \cup\text{hypo} (\text{ENF}_{p_2}))
\end{align*}
\end{itemize}
\end{lemma}
\textit{Proof:} See Appendix \ref{monotonic}.
}

\section{Tropical Polynomial Division Definition and Existence}
\label{Sec:Theory}
We first define the tropical polynomial division. 
\begin{definition}
Let $p,d$ be tropical polynomials. We define the quotient and the remainder of the division of $p$ by $d$. 
\begin{itemize}
\item[(a)] A tropical polynomial $q$ is the quotient if 
\begin{equation}
p(\bld x)\geq q(\bld x)+d(\bld x), \text{~~~~~~for all } \bld x\in \mathbb R^n, \label{QuotDef}
\end{equation}
and $q$ is the maximum polynomial satisfying this inequality. Particularly, for any tropical polynomial $\tilde q$, such that $p(\bld x)\geq \tilde q(\bld x)+d(\bld x)$, for all $\bld x\in\mathbb R^n$, it holds $\tilde q(\bld x)\leq  q(\bld x),$ for all $ \bld x\in \mathbb R^n$.
\item[(b)] A tropical polynomial $r$ is the remainder of the division of $p$ by $d$  with quotient $q$ if 
\begin{equation}
\label{remEq}
p(\bld x)= (q(\bld x)+d(\bld x))\vee r(\bld x), \text{~~~~~~for all } \bld x\in \mathbb R^n,
\end{equation}
and $r$ is the minimum tropical polynomial satisfying this equality. Particularly, for any tropical polynomial $\tilde r$, such that $p(\bld x)= (q(\bld x)+d(\bld x))\vee \tilde r(\bld x)$, it holds $\tilde r(\bld x)\geq  r(\bld x),$ for all $ \bld x\in \mathbb R^n$.
\end{itemize}
\end{definition}

Observe that the set of tropical polynomials $q$ satisfying \eqref{QuotDef} is closed under the pointwise maximum. That is, assume that
 $q_1,q_2$ are tropical polynomials such that $p(\bld x) \geq d(\bld x)+q_1(\bld x)$ and $p(\bld x) \geq d(\bld x)+q_2(\bld x)$, then  it holds $p(\bld x) \geq d(\bld x)+(q_1(\bld x)\vee q_2(\bld x))$. 
This property motivates us to determine the monomials $$q_M^{\bld a,b}(\bld x) = \bld a^T\bld x+b,$$
that satisfy \eqref{QuotDef}.  For each monomial coefficient $\bld a$, define the maximum value of $b$ that satisfy $p(\bld x)\geq d(\bld x)+q_M^{\bld a,b}(\bld x)$ as 
\begin{align}
l(\bld a) = \sup \{b: q_M^{\bld a,b}(\bld x)+d(\bld x)\leq p(\bld x), \text{for all } x\in \mathbb R^n\}. \label{l_def}
\end{align}
Note that $l(\bld a)$ can be written as
\begin{align*}
l(\bld a) &= \sup \{b: b\leq p(\bld x)-d(\bld x)-\bld a^T \bld x, \text{for all } x\in \mathbb R^n\}\\
&=\inf_{\bld x\in \mathbb R^n} \{p(\bld x)-d(\bld x)-\bld a^T \bld x\}.
\end{align*}
Denoting by $f(\bld x)$ the difference $p(\bld x)-d(\bld x)$, we have 
\begin{align*}
l(\bld a) &=-f^\star(\bld x),
\end{align*}
where $f^\star$ is the convex conjugate of $f$. 
A candidate for the quotient is 
\begin{align*}
q(\bld x)  = \sup_{\bld a \in \mathbb R^n} \{\bld a ^T\bld x+l(\bld a) \}.
\end{align*}
Note that $q$ is the bi-conjugate of $f$. Indeed
\begin{align}
q(\bld x)  = \sup_{\bld a \in \mathbb R^n} \{\bld a ^T\bld x-f^\star ({\bld a}) \}=f^{\star\star}(\bld x ). \label{Quot_formula}
\end{align}
 
The following proposition shows that $q(\bld x)$ is indeed the quotient of the division. 

\begin{proposition}
For any pair of tropical polynomials $p,d$ there is a quotient-remainder pair. The quotient is given by \eqref{Quot_formula}. Furthermore, the polynomial functions of the quotient and remainder are unique.
\label{ExistProp}
\end{proposition}
\textit{Proof} See Appendix \ref{Proof1}.

\begin{remark}
The use of the convex conjugate is closely related to the slope transform  \cite{maragos1994morphological,maragos1995slope,heijmans1997lattice}. The slope transform was used to derive an analog of frequency response of max-plus and more generally morhological systems.
\end{remark}

\begin{example}[Tropical Polynomial Division in 1D]
\label{Ex1d}
Let $p(x) = \max(-2x-1,1,x+1,3x-3)$ and $d(x)=\max(x,2x-1)$. Equivalently, the polynomial functions $p$ and $d$ are written as
$$p(x)=\begin{cases} -2x-1,~\text{if~} x\leq-1\\1,~~~~~~~~~\text{if~} -1<x\leq0\\x+1,~ ~~~\text{if~} 0<x\leq2 \\ 3x-3,~~~\text{if~} x\geq 2
 \end{cases},$$ $$ d(x)=\begin{cases} x,~~~~~~~~\text{if~} x\leq1\\2x-1,~~\text{if~} x>1
 \end{cases}.$$
 Thus,
 $$f(x)=p(x)-d(x)=\begin{cases} -3x-1,~\text{if~} x\leq-1\\
 1-x,~~~~~\text{if~} -1<x\leq0\\
 1,~~~~~~~ ~~~\text{if~} 0<x\leq1 \\
 -x+2,~~~\text{if~} 1<x\leq2 \\ 
 x-2,~~~~~\text{if~} x\geq 2
 \end{cases}$$
The plots of $p,d$ and $f$ are given in Figure \ref{SINgleDimEx}. The difference $f$ is not convex since the slopes $-3,-1,0,-1,1$ are not increasing. The quotient $q$ is given as the convex bi-conjugate of $f$, that is the largest convex function which is less than or equal to $f$, for all $x$. Thus, the quotient is given $q(x)=\max(-3x+1,1-x,-0.5x+1, x-2)$. The sum $d(x)+q(x)$ is equal to $p(x)$, for all $x$, except $x\in(0,2)$. Thus, $r(x)=x+1$, that is, the form of $p(x)$, for $x\in(0,2)$.

\begin{figure}
\centering
\includegraphics[width=0.5\textwidth]{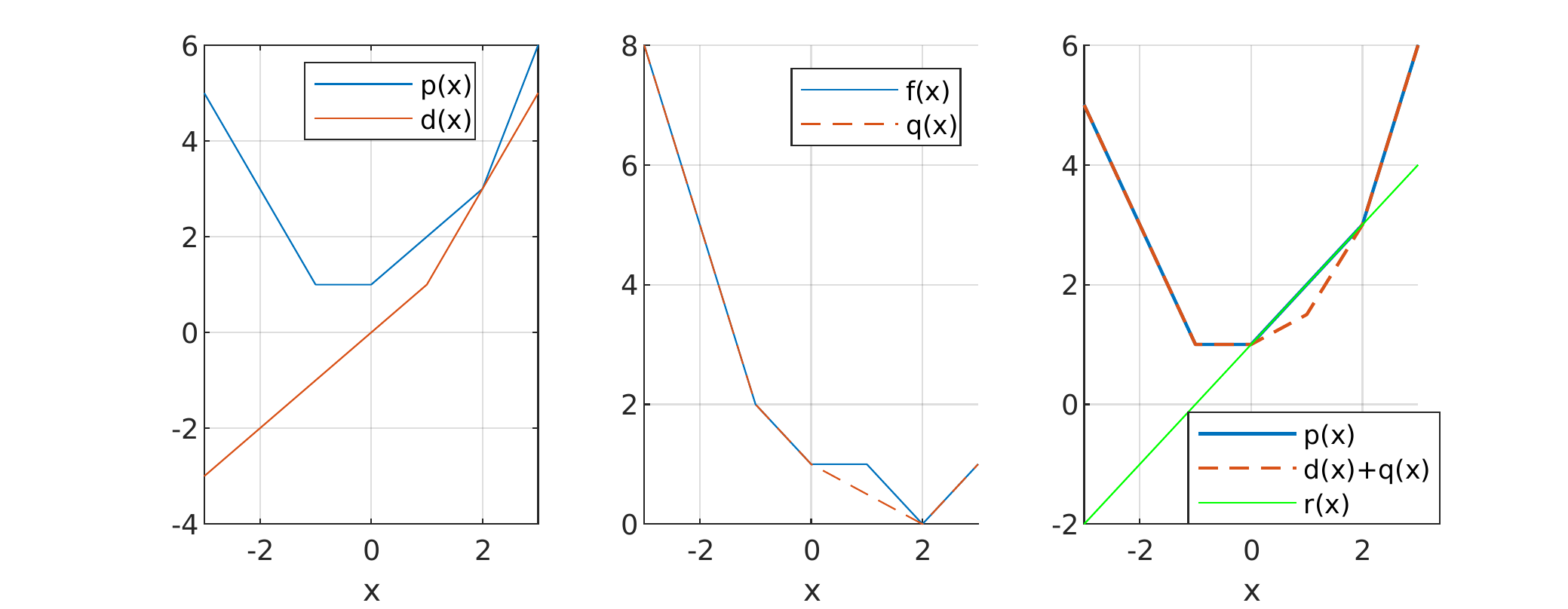}
\caption{\textit{This figure refers to Example \ref{Ex1d}. The left plot presents the dividend $p$ and the divisor $d$. The middle plot their difference $f$ and the quotient $q$. The right part shows the dividend $p$, its approximation $d+q$ and the remainder $r$. }}
\label{SINgleDimEx}
\end{figure}
\end{example}

We call the division \emph{nontrivial} if $q(\bld x)>-\infty$ and \emph{effective} if $r(\bld x) \neq p(\bld x)$, for some $\bld x\in \mathbb R^n$. It is easy to see that an effective division is also nontrivial. As shown in the previous example, these notions are not equivalent.

\begin{proposition}
\label{EffectiveProp}
Assume that $p$ and $d$ are tropical polynomials. If the division of $p$ by $d$ has quotient $q$ and remainder $r$. Then:
\begin{itemize}
\item [(a)]  The division of $r$ by $d$ is not effective.
\item[(b)] The division of $p$ by $q$ has quotient $d$ and remainder $r$.
\end{itemize}
\end{proposition}
\textit{Proof} See Appendix \ref{Proof2}.

\begin{remark}
For the Euclidean division of a positive integer  $p$ by a positive integer $d$ with quotient $q$ and remainder $r$, the notion of nontrivial division corresponds to $q\neq0$ and effective division to $r< p$.   For the division of positive integers, we observe that these notions are equivalent. Furthermore,  the converse of (a) is true. Particularly, if for some $\bar q,\bar r \in \mathbb N$, it holds $p=d\bar q+\bar r$ and the division of $r$ by $d$ is not effective, then $\bar q$ is the quotient and $\bar r$ the remainder. 

On the other hand, for tropical polynomials, the converse of (a) is not true. Particularly, it is possible that for a pair of tropical polynomials $p,d$ there is another pair $q,r$ such that \eqref{remEq} is satisfied, the division of $r$ by $d$ is not effective (or even  is trivial), but $q,r$ is not a quotient-remainder pair. For example consider $p(x)=\max(3x,-3x), d(x) =\max(2x,-2x) $ and $q(x)=\max(0,x), r(x) =\max(0, -3x)$. It is easy to see that the division of $r$ by $d$ is trivial.  
\end{remark}

Consider the set 
\begin{align}
\label{SetC}
C = \{\bld c\in \mathbb R^n: \text{Newt}(\bld c^T\bld x+d(\bld x))\subset  \text{Newt}(p(\bld x))\}.
\end{align}
This set appears also \cite{smyrnis2019tropical,smyrnis2020maxpolynomial, smyrnis2020multiclass} for the case of discrete coefficients.
It is not difficult to see that $C$ is a convex polytope. 
The following proposition shows that the division is non-trivial if and only if $C\neq0$.

\begin{proposition} Let $C$ be the set defined in \eqref{SetC}. Then,
\label{Prop_setC}
\begin{itemize}
\item[(a)] $C$ is equal to the domain of function $l(\cdot)$ (given by \eqref{l_def}). That is,
$$C=\text{\textbf {dom}} (l(\bld a))=\{\bld a \in \mathbb R^n: l(\bld a)>-\infty\}.$$
\item[(b)] The division is non-trivial if and only if $C\neq \emptyset$
\item[(c)]  It the quotient has the form  $q(\bld x) = \bigvee_{i=1}^ {m_q} (\hat {\bld a}_i^T\bld x+\hat b_i)$, then $\hat {\bld a}_i\in C$, for all $i$.
\end{itemize}
\end{proposition}
\textit{Proof} See Appendix \ref{Proof3}.

\begin{corollary}
\label{Corollary_DimRed}
Assume that  the division of a tropical polynomial $p(\bld x)= \bigvee_{i=1}^{m_p} (\boldsymbol{a}_i^T\boldsymbol{x}+b_i)$ by another tropical polynomial $d(\bld x)= \bigvee_{i=1}^{m_d} (\tilde{\boldsymbol{a}}_i^T\boldsymbol{x}+\tilde b_i)$ is nontrivial. Then
\begin{itemize}
\item[(i)] It holds
 $$ \text{span}\{\tilde {\bld  a}_1,\dots, \tilde {\bld  a}_{m_d}\}\subset\text{span}\{\bld a_1,\dots,\bld a_{m_p}\}. $$
\item[(ii)] The quotient $q(\bld x)= \bigvee_{i=1}^{m_q} (\hat{\boldsymbol{a}}_i^T\boldsymbol{x}+\hat b_i)$ is such that
$$ \text{span}\{\hat{\bld  a}_1,\dots, \hat{\bld  a}_{m_q}\}\subset\text{span}\{\bld a_1,\dots,\bld a_{m_p}\}. $$
\end{itemize}
\end{corollary}
\textit{Proof:} See Appendix \ref{Corollary_DimRedProof}.

\begin{remark}
\label{RedDimRem}
Corollary \ref{Corollary_DimRed} can be used to simplify the division of two tropical polynomials. Particularly, assume that   $$\text{span}\{\tilde {\bld  a}_1,\dots, \tilde {\bld  a}_{m_d}\}\subset\text{span}\{\bld a_1,\dots,\bld a_{m_p}\}\subsetneq \mathbb R^d.$$ 
If $\bld Q$ is a matrix the columns of which represent an orthonormal  basis of the subspace $\text{span}\{\bld a_1,\dots,\bld a_{m_p}\}$, then $\bld a_i$ can be expressed in terms of a reduced dimension vector $\bld a_i^r$ as $\bld a_i=\bld Q\bld a_i^r$, and $\tilde{\bld a}_i$  similarly in terms of $\tilde {\bld  a}_i^r$ as $\tilde {\bld  a}_i=\bld Q\tilde {\bld  a}_i^r$. Then, the division of polynomials $p$ and $d$ can be expressed in terms of reduced dimension polynomials $p^r,d^r$  as $p(\bld x)=p^r(\bld Q^T\bld x)=p^r(\tilde{\bld x}) ,d=d^r(\bld Q^T\bld x)=d^r(\tilde{\bld x})$, where the dimension of $\tilde{\bld x}$ is equal to the rank of $\bld Q$.

This observation is certainly useful when the number of terms of the dividend $m_p$ is less than the space dimension $d$. It will be also used in Section \ref{Sec:Composite}.
\end{remark}

\section{An Exact Algorithm For Tropical Division}
\label{Sec:ExactAlg}

We now present an algorithm to compute the quotient and remainder of a division. The algorithm uses polyhedral computations. Particularly, we use
\eqref{Quot_formula}, and its equivalent in terms of the epigraphs (see \eqref{epigraph_ch} in the appendix) to compute $q$.

 The input of the algorithm is a pair of tropical polynomials $$p(\bld x)= \bigvee_{i=1}^{m_p}(\bld a_i^T\bld x+b_i),~~~d(\bld x)=\bigvee_{j=1}^{m_d}(\tilde {\bld a}_j^T\bld x+\tilde b_j).$$
The first step is to compute a partition of $\mathbb R^n$ in  polyhedra on which $f(\bld x)=p(\bld x)-d(\bld x)$ is linear. To do so, for each $i,j$, consider the polyhedron on which the terms $i,j$ attain the maximum in $p$ and $d$ respectively.
\begin{align*} P_{i,j}&=\{x\in\mathbb R^n: \bld a_i^T\bld x+b_i\geq \bld a_{i'}^T\bld x+b_{i'},\\& \tilde {\bld a}_j^T\bld x+\tilde b_j\geq \tilde  {\bld a}_{j'}^T\bld x+\tilde  b_{j'},i'=1,\dots,m_p,j'=1,\dots,m_d\}.
\end{align*}
Polyhedron $P_{i,j}$ can be compactly written in terms of a matrix $\bld A^{i,j}$ and a vector $\bld  b^{i,j}$ as
$$P_{i,j} = \{\bld x\in \mathbb R^n: \bld A^{i,j}\bld x\geq \bld  b^{i,j}\}.$$
Note that some $P_{i,j}$'s can be empty.

The second step is to compute the polyhedra that represent the epigraph of $f(\bld x)=p(\bld x)-d(\bld x)$ for each $P_{i,j}$. The epigraphs are given by: 
\begin{align*}
E_{i,j} &= \{(\bld x, z)\in \mathbb R^{n+1}:\bld  A^{i,j}\bld x\geq \bld  b^{i,j}, \\&~~~~~~z\geq(\bld a_i -\tilde  {\bld a}_j)^T\bld x+b_i-\tilde  b_j\}. 
\end{align*}

For each polyhedron $E_{i,j}$, we compute  the corresponding $\mathcal V-$representation, consisting of a set of vertices $V_{i,j}$ and a  set of rays $R_{i,j}$, satisfying \eqref{resol_equ}. Note that the epigraph of $f$ is given by $\cup_{i,j} E_{i,j}$.

 The epigraph of the quotient $q$ is given by the closed convex hull of the epigraph of $f$ (see \eqref{epigraph_ch}  in the appendix). 
Next, we consider the union of vertices  $V=\cup_{i,j}V_{i,j}$, and rays $R=\cup_{i,j}R_{i,j}$, and consider the polyhedron $ E$ generated by $V$ and $R$ (as in \eqref{resol_equ}). Due to Proposition  \ref{ExistProp}, $E$ is the epigraph of the quotient $q(\bld x)$. 
Then, compute the $\mathcal H-$representation of $ E$
\begin{align}
\label{E_formula}
 E=\{(\bld x, z)\in \mathbb R^{n+1} :  [\bld A^{E,\bld x}~ \bld a^{E,z}][\bld x^T~ z]^T\geq \bld b^E\},
\end{align}
for appropriate matrix $\bld A^{E,\bld x}$ and  vectors $\bld a^{E,z}, \bld b^E$. Assume that $E\neq \mathbb R^n$. We will prove in Proposition \ref{Alg_prop}  that the components of $\bld a^{E,z} $ are positive. Thus,  
\begin{align*}
 E=&\left\{(\bld x, z)\in \mathbb R^{n+1} :    z\geq \bld -\frac{1}{[\bld a^{E,z}]_l}[\bld A^{E,\bld x}]_l \bld x +\right.\\&~~~~~\left.+[\bld b^E]_l/[\bld a^{E,z}]_l,~l=1,\dots,L\right\},
 \end{align*}
where $L$ is the number of rows of $\bld A^{E,\bld x}$. Therefore,
\begin{align}
\label{q_formula}
q(\bld x) = \bigvee_{l=1}^L\left(\bld -\frac{1}{[\bld a^{E,z}]_l}[\bld A^{E,\bld x}]_l \bld x +[\bld b^E]_l/[\bld a^{E,z}]_l\right).
\end{align}

\begin{figure}
\centering
\includegraphics[width=0.4\textwidth]{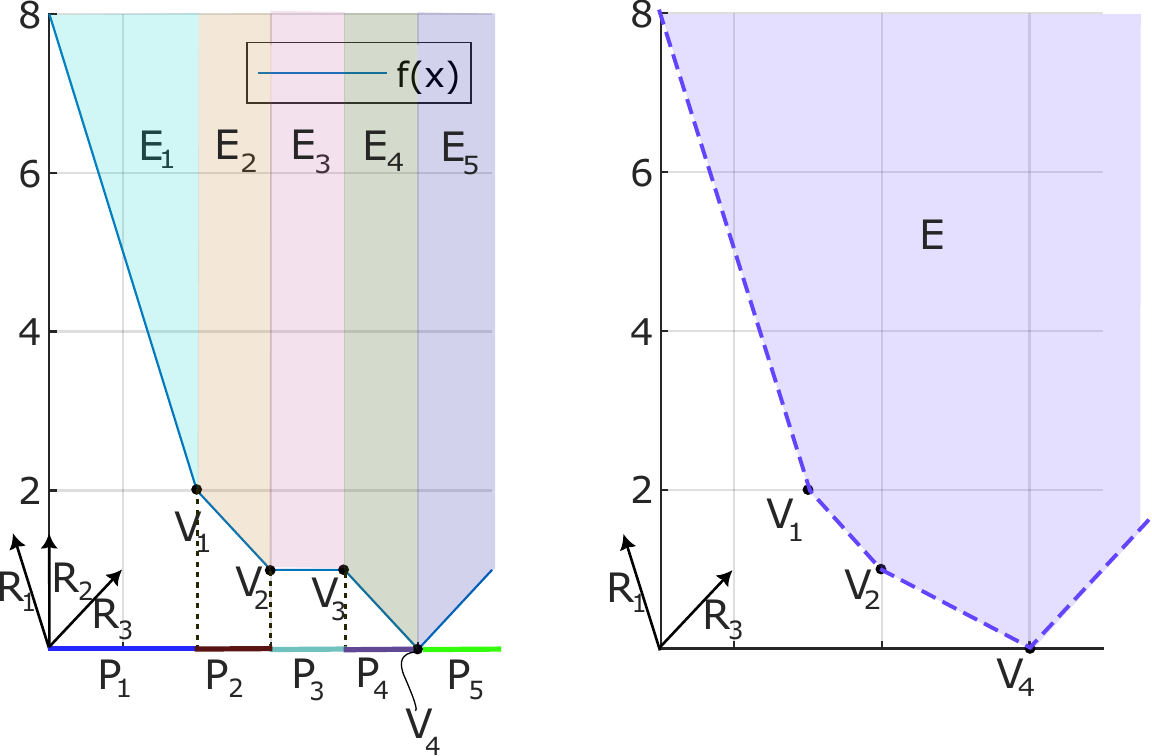}
\caption{\textit{Illustration of the algorithm applied to Example \ref{Ex1d}. Function $f(x) = p(x)-d(x)$ is linear on each of the subsets $P_1,\dots,P_5$. The epigraphs are generated as follows: $E_1$ by vertex $V_1$ and rays $R_1$ and $R_2$, epigraph $E_2$ by vertices $V_1, V_2$ and ray $R_2$, epigraph $E_3$ by vertices $V_2, V_3$ and ray $R_2$,  epigraph $E_4$ by vertices $V_3, V_4$ and ray $R_2$, and finally  epigraph $E_5$ by vertex $V_4$ and rays $R_2,R_3$. Their closed convex hull $E$ is generated by vertices $V_1,V_2,V_4$ and rays $R_1,R_3$. The quotient $q(x)$ is computed by the $\mathcal H-$representation of $E$.}}
\label{SINgleDimEx_b}
\end{figure}

The procedure is illustrated in Figure \ref{SINgleDimEx_b}.

Having computed the quotient, we formulate the sum
$\tilde p (\bld x)= d(\bld x) +q(\bld x)$. For each $P_{i,j}$ we choose a point $\bld x_{i,j}$ in its relative interior, for example 
\begin{align}
\label{InterPoints}
\bld x_{i,j}=\frac{1}{|V_{i,j}|}\sum_{v \in V_{i,j}} v+\sum_{w\in R_{i,j}}w.
\end{align}
 Let $I$ be the set of indices $i$ such that there is an index $j$ satisfying $p(\bld x_{i,j})>\tilde p(\bld x_{i,j})$. Then,
\begin{align}
r(\bld x) = \bigvee_{i\in I} (\bld a_i^T \bld x+b_i).
\label{RemComp}
 \end{align}
The computation of the quotient and the remainder is summarized in Algorithm \ref{alg:ExactPolyDivi}.

\begin{algorithm}
    \caption{Exact Polynomial Division}
    \label{alg:ExactPolyDivi}
    \begin{algorithmic}
 \item[0:] Input: Tropical Polynomials $p(\bld x)$ and $d(\bld x)$
 \item[1:] Compute a partition of $\mathbb R^n$ into  polyhedra $P_{i,j}$, $i=1,\dots,m_p$, $j=1,\dots,m_d$, and their interior points $\bld x_{i,j}$ according to \eqref{InterPoints}
 \item[2:] Compute the epigraphs $E_{i,j}$
 \item[3:] For each epigraph, compute the resolution into a set of vertices $V_{i,j}$ and a  set of rays $R_{i,j}$, using a representation conversion algorithm 
 \item[4:] Compute the $\mathcal H-$representation of the polyhedron generated by the union of the set of vertices $V=\cup_{i,j}V_{i,j}$, and the union of the set of rays $R=\cup_{i,j}R_{i,j}$, in the form \eqref{E_formula}
 \item[5:] If $L>0$ then $q(\bld x)$ is given by \eqref{q_formula}. Otherwise, $q(\bld x)=-\infty$.
 \item[6:]   Compute the set of indices $i$ such that there is a $j$ with $p(\bld x_{i,j})-d(\bld x_{i,j})-q(\bld x_{i,j})>0$. Compute $r(\bld x)$ according to 
\eqref{RemComp}
 \item[7:] Return $q(\bld x)$, $r(\bld x)$
    \end{algorithmic}
\end{algorithm}

\begin{proposition}
\label{Alg_prop}
If $E\neq \mathbb R^n$, then the output of the algorithm is indeed the quotient and the remainder of the division. If $E=\mathbb R^n$, then the quotient is $q(\bld x)=-\infty$, for all $x\in\mathbb R^n$ and the remainder is equal to the dividend, i.e.,  $r=p$.
\end{proposition}
\textit{Proof} See Appendix \ref{Proof4}.

\begin{remark}
The algorithm involves first computing the resolution of a number of polyhedra and then the computation of the  $\mathcal H-$representation of the convex hull of their union. The computational complexity of these operations is exponential on their input size (e.g. \cite{fukuda2004frequently}). Therefore, we expect that the algorithm will be usable only for small examples.
\end{remark}
 
 \begin{example}
\label{ExactExample}
 Let us divide the tropical polynomial $p(x,y) = \max(0,3x+3y,6x),$ by  $d(x,y) = \max(x,x+y,2x+y)$, using Algorithm \ref{alg:ExactPolyDivi}. The quotient is 
\begin{align*}
q(x,y ) = \max(1.5x +1.5y, 3x,0),
\end{align*}
and $d(x)+q(x)$ is given by $
 \max(x,x+y,2.5x+2.5y,3.5x+2.5y,5x+y,4x)$.

 The linearity regions of $f(x,y)$ and $q(x)$ are shown in Figure \ref{TwoDimEx}. Furthermore, Figure  \ref{TwoDimEx} illustrates the Newton polygons for $p,d$ and $d+q$.
\begin{figure}
\centering
\includegraphics[width=0.5\textwidth]{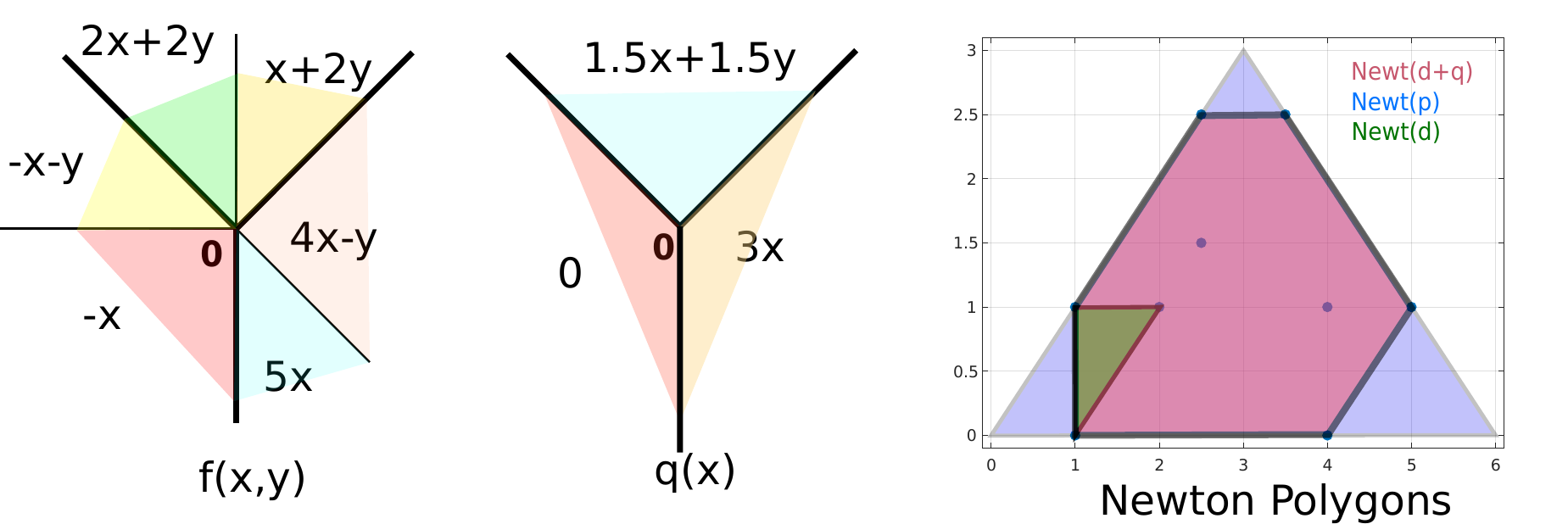}
\caption{\textit{The linear regions of $f(x,y)$ and $q(x,y)$. The two functions coincide on the bold half-lines. The figure also illustrates the Newton polygons of $p(x,y)$, $d(x,y)$, and $d(x,y)+q(x,y)$.}}
\label{TwoDimEx}
\end{figure}
 \end{example}

\section{Approximate  Algorithms For Tropical  Division}

\label{Sec:ApproxAlg}

We now present an approximate algorithm for computing the quotient of a tropical division,  assuming that the quotient has a predefined maximum number of terms $\tilde m_q$. The algorithm mimics the procedure developed in \cite{magnani2009convex}.

The quotient has the form $q(\bld x) = \bigvee_{i=1}^ {m_q} (\hat {\bld a}_i^T\bld x+\hat b_i)$. From the proof of Proposition \ref{ExistProp} we know that $q(\bld x)$ is the maximum tropical polynomial function, all the terms of which satisfy $ \hat {\bld a}_i^T\bld x+\hat b_i\leq f(\bld x)$, for all $\bld x\in \mathbb R^n$. We will utilize this property  to formulate an optimization problem on a set of sample points $\bld x_1,\dots,\bld x_N \in\mathbb R^n$.
Given the set of sample points and the corresponding set of the values of function  $f(\cdot)$, i.e., $f(\bld x_1),\dots,f(\bld x_N) $, the optimization problem is formulated as:
\begin{align}
\begin{aligned}
& \underset{ \hat {\bld a}_i,\hat b_i, i=1,\dots,\tilde m_q}{\text{maximize}}
& & \sum_{j=1}^{N}\bigvee_{i=1}^{\tilde m_q} (\hat {\bld a}_i^T\bld x_j+\hat b_i)\\
& \text{subject to}
& & \hat {\bld a}_i^T\bld x_j+\hat b_i\leq f(\bld x_j),~~\forall i,j\\
& & &\hat {\bld a}_i\in C,~~i=1,\dots,\tilde m_q
\end{aligned}~~,
\label{Prob1}
\end{align}
where $C$ is the set defined in \eqref{SetC}. 
Since $C$ is a polytope, the last constraint in  \eqref{Prob2} can be written as a set of linear inequalities. The details are presented in subsection \ref{C_set_sect}.

The algorithm alternates between two phases. Phase 1 partitions the data. Particularly, assuming a set of solutions $ (\hat {\bld a}_i,b_i)$,   $i=1,\dots,\tilde m_q$, we partition the samples $\bld x_j$ into sets $I_1,\dots, I_{\tilde m_q}$ with $j\in I_i$, if \sloppy
\begin{align}
\label{Partition}
\hat {\bld a}_i^T\bld x_j+\hat b_i\geq  \hat {\bld a}_{i'}^T\bld x_j+\hat b_{i'},~\text{  for all } i'.
\end{align}

Phase 2, solves the linear optimization problem: 
\begin{align}
\begin{aligned}
& \underset{ \hat {\bld a}_i,b_i, i=1,\dots,\tilde m_q}{\text{maximize}}
& & \sum_{i=1}^{\tilde m_q}\sum_{j\in I_i}  \hat {\bld a}_i^T\bld x_j+\hat b_i\\
& \text{subject to}
& & \hat {\bld a}_i^T\bld x_j+\hat b_i\leq f(\bld x_j),~~\forall i,j\\
& & &\hat {\bld a}_i\in C,~~i=1,\dots,\tilde m_q
\end{aligned}~~.
\label{Prob2}
\end{align}

We then simplify \eqref{Prob2} in two ways. First, it may contain redundant inequalities. To remove the redundant inequalities we compute the  lower convex hull of the set $\{(\bld x_j,f(\bld x_j)):j=1,\dots,N\}$ with respect to its last component, and denote by $J$ the set of indices $j$, for which $(\bld x_j,f(\bld x_j)$ belongs to the lower convex hull. 

Second, problem \eqref{Prob2} decomposes into a set of $ \tilde m_q$  linear programs
\begin{align}
\begin{aligned}
& \underset{ \hat {\bld a}_i,b_i}{\text{maximize}}
& & \bld s_i^T\hat {\bld a}_i+N_i \hat b_i\\
& \text{subject to}
& & [ (\bld x_j)^T ~1]   \left [ \begin{matrix} \hat {\bld a}_i \\ \hat b_i  \end{matrix}     \right] \leq f(\bld x_j),~~j\in J\\
& & &\hat {\bld a}_i\in C
\end{aligned}~~,
\label{Prob3}
\end{align}
where $\bld s_i=\sum_{j\in I_i} \bld x_j$, $N_i = |I_i|$.

The computations are summarized in Algorithm \ref{alg:ApproxPolyDivi}.

\begin{algorithm}
    \caption{Approximate Polynomial Division}
    \label{alg:ApproxPolyDivi}
    \begin{algorithmic}
 \item[0:] Input: Tropical Polynomials $p(\bld x)$, $d(\bld x)$,  the degree of the approximate quotient $\tilde m_q$, the number of sample points $N$, the maximum number  of iterations
 \item[1:] Compute the inequalities corresponding to set $C$ using the formulas in Appendix \ref{C_set_sect}
 \item[2:] Sample the points $\bld x_j$, for $j=1,\dots,N$, and compute the values of $f(\bld x_j)=p(\bld x_j)-d(\bld x_j)$ 
 \item[3:] Compute the    lower convex hull of the set $\{(\bld x_j,f(\bld x_j)):j=1,\dots,N\}$, and use the samples belonging to it to formulate the first constraint in  \eqref{Prob2}
 \item[4:] Initialize the partition $I_1,\dots,I_{\tilde m_q}$
 \item[5:]  Solve Problems \eqref{Prob3}, for all $i$
 \item[6:]  Compute the partitions $I_i$ that satisfy \eqref{Partition}
 \item[7:]  If the number of iterations has not exceeded the maximum number of iterations go to Step 5 
 \item[8:] Return $\hat q(\bld x) = \bigvee_{i=1}^{\tilde m_q} \hat {\bld a}_i^T\bld x+\hat b_i $,  
    \end{algorithmic}
\end{algorithm}

We then examine the quality of approximation of $p(\bld x)$ by $d(\bld x)+q_t(\bld x)$ on the sample set given by
\begin{equation}
e(t) = \sum_{j=1}^N [p(\bld x_j) -d(\bld x_j)-q_t(\bld x_j)],
\end{equation}
where $q_t$ is the quotient computed after $t$ steps of the algorithm. The following proposition shows that the quality of approximation improves as the number of steps increases. 

\begin{proposition}
\label{MonotProp}
The quantity $e(t)$ is nonnegative and non-increasing with respect to $t$.
\end{proposition}
 \textit{Proof} See Appendix \ref{Proof5}.
\begin{remark} In practice some issues with Algorithm 1 may arise. Some of the classes $I_i$ may end up empty. In this case, we may split another existing class randomly. Furthermore, we expect the algorithm to converge to a local optimum. We may use a multiple start method to avoid bad local optima. 
\end{remark}

\begin{remark} 
Let us comment about the relationship between Algorithm \ref{alg:ApproxPolyDivi} and
\cite{magnani2009convex}. Inspired by \cite{magnani2009convex}, Algorithm \ref{alg:ApproxPolyDivi} partitions the data in several clusters.  The major difference is that in Algorithm \ref{alg:ApproxPolyDivi}  we use the sharper lower linear approximation of each class of data subject to the constraint that this approximation does not exceed any sample point of any class. Thus, instead of alternating between clustering and linear regression, we alternate between clustering and linear programming.

Another related work is \cite{kim2010convex} which studies the problem of  computing a convex under-approximation  for a given dataset by a piecewise linear function. The algorithm uses an iterative linear programming technique and has two phases. At phase one of each iteration, it considers a subset of the data set and minimizes the  maximum approximation error for this subset, using a piecewise linear function with a number of terms equal to the iteration count. At phase two, it adds to this subset the data point that maximizes the approximation error. Then, it moves to the next iteration. 
\end{remark}

\subsection{The Linear Constraints for Set $C$}
\label{C_set_sect}

We next compute a set of linear inequalities describing set $C$. Let the extreme points of $\text{Newt}(p) $ be $\bld a_{e,1},\dots,\bld a_{e,k}$.
 Then,  $\hat{\bld a}\in C$ if and only if $\hat{\bld a}+\text{Newt}(d(\bld x))\subset  \text{Newt}(p(\bld x)).$ That is,
$$\hat {\bld a } + \tilde {\bld a }_j \in \text{conv}\{\bld a_{e,1},\dots,\bld a_{e,k}\},$$
for all $j$. Equivalently if there are auxiliary variables $\lambda_{j,l}$,  such that
\begin{align*}
\hat{\bld a} +\tilde {\bld a}_j&= \sum_{l=1}^k \bld a_{e,l} \lambda_{j,l},  ~~~
\lambda_{j,l}\geq 0,~~~
\sum_{l=1}^k \lambda_{j,l}=1
\end{align*}
for all $j=1,\dots, m_d$, $l=1,\dots,k$.

Problem \eqref{Prob3} becomes:
\begin{align}
\begin{aligned}
& \underset{ \hat {\bld a}_i,b_i,\lambda}{\text{maximize}}
& & \bld s_i^T\hat {\bld a}_i+N_i \hat b_i\\
& \text{subject to}
& & [ \bld x_j^T ~1]   \left [ \begin{matrix} \hat {\bld a}_i \\ \hat b_i  \end{matrix}     \right] \leq f(\bld x_j),~~j\in J\\
& & &\hat{\bld a}_i +\tilde {\bld a}_j= \sum_{l=1}^k \bld a_{e,l} \lambda_{j,l},  ~~\\
& & &\lambda_{j,l}\geq 0\\
& & &\sum_{l=1}^k \lambda_{j,l}=1
\end{aligned}~~.
\label{Prob4}
\end{align}

\mycomment{

\subsection{An Alternative Approximate Algorithm}
In this section, we propose an alternative approximate algorithm utilizing the extended Newton function to represent inequality $d(\bld x)+q(\bld x)\leq p(\bld x)$. 
This Inequality 
is equivalent to $\text{ENF}_{q+d}(\bld a)\leq \text{ENF}_{p}(\bld a)$, 
for all $\bld a\in\mathbb R^n$ (see Appendix \ref{monotonic}). 

Based on this formulation, we may apply Algorithm \ref{alg:ApproxPolyDivi}, but instead of  Problem \eqref{Prob3} to solve
\begin{align}
\begin{aligned}
& \underset{ \hat {\bld a}_i,b_i,\lambda}{\text{maximize}}
& & \bld s_i^T\hat {\bld a}_i+N_i \hat b_i\\
& \text{subject to}
& &  \hat{\bld a} +\tilde {\bld a}_j= \sum_{l=1}^k \bld a_{l} \lambda_{j,l},  \\
& & &  \hat{  b} +\tilde {\bld b}_j\leq \sum_{l=1}^k   b_{l} \lambda_{j,l},  \\
& & &\lambda_{j,l}\geq 0\\
& & &\sum_{l=1}^k \lambda_{j,l}=1
\end{aligned}~~.
\label{Prob5}
\end{align}

We refer to the modified algorithm as Algorithm \ref{alg:ApproxPolyDivi}-ENF.

\begin{proposition}
Any tropical polynomial $q(\bld x)$, obtained using Algorithm \ref{alg:ApproxPolyDivi}-ENF satisfies $d(\bld x)+q(\bld x)\leq p(\bld x)$.
\label{PropENFalg}
\end{proposition}
\textit{Proof:} See Appendix \ref{ProofPropENFalg}. 

}

%\mycomment
{
\section{Division for Composite Polynomials}
\label{Sec:Composite}

In this section, we present some analytical results and some algorithms for dividing tropical polynomials that are sums or maxima of simpler ones.  

\subsection{General Properties}

We will use $Q(p,d)$ and $R(p,d)$ to denote the quotient and remainder of the division of a tropical polynomial $p$ by another tropical polynomial $d$. 

\begin{proposition}
\label{IneqProp}
The following inequalities hold
\begin{align}
Q(p_1+p_2,d)&\geq Q(p_1,d)+Q(p_2,d)+d, \label{Ineq1}\\
Q(p_1\vee p_2,d)&\geq Q(p_1,d)\vee Q(p_2,d),\label{Ineq2}\\
Q(p,d_1+d_2)&\geq Q(p,d_1)+ Q(p,d_2)-p,\label{Ineq3}\\
Q(p,d_1\vee d_2)&\geq Q(p,d_1)\wedge Q(p,d_2).\label{Ineq4}
\end{align}
Furthermore, if $R(p_1,d)=R(p_2,d)=-\infty$, then \eqref{Ineq1},\eqref{Ineq2},\eqref{Ineq3} hold as equalities.
Finally, if $s$ is a monomial, i.e., it has the form $s(\bld x)=\bld a ^T\bld x +b$, then 
\begin{align}
Q(p+s,d)&= Q(p,d)+s=Q(p,d-s), \label{Ineq5}
\end{align}
\end{proposition}
\textit{Proof} See Appendix \ref{Proof6}.

\begin{remark}
Assume that polynomial $d$ is fixed. Inequality \eqref{Ineq2} implies that function $Q(\cdot,d)$ is nondecreasing. If  $d(\bld x)\geq 0$, for all $\bld x\in\mathbb R^n$, then function $Q(\cdot,d)$ is also super-additive. 
\end{remark}

In the following we call a tropical polynomial described as a summation of other tropical polynomials as composite. Furthermore, we refer to a tropical polynomial in the form \ref{Pdef} as simple\footnote{These notions refer to the representation of the tropical polynomials, and not to whether or not they can be written as a sum of other polynomials. That is, it is possible that a tropical polynomial with a simple representation (in the form \eqref{Pdef}), can be factorized and written in a composite form.}.

}
\subsection{Dividing a Composite Polynomial by a Simple}
\label{Sec:Composite}

In the following we propose  algorithms to approximately   divide a composite polynomial $p(\bld x)=\sum_{\nu=1}^N p^\nu(\bld x)$ or $p(\bld x)=\bigvee_{\nu=1}^N p^\nu(\bld x)$  by another polynomial $d(\bld x)$. The computation of the value of the composite polynomial $p(\bld x)$ is trivial, provided the values of the simpler polynomials $p^\nu(\bld x)$. Thus, the only difficulty in order to apply Algorithm \ref{alg:ApproxPolyDivi} is the computation of the Newton polytope $\text{Newt}(p).$ 

It is convenient to consider  the Newton polytopes $\text{Newt}(p^\nu)$ in terms of an extended description. Particularly, assume matrices $\bld {A}^\nu,\bld  F^\nu$, and vectors $\bld \beta^\nu$ of appropriate dimensions, such that \sloppy
\begin{equation}
\text{Newt}(p^\nu)  = \{ \bld a\in \mathbb R^n:\bld a=\bld F^\nu \bld \alpha^\nu,  \bld {A}^\nu \bld \alpha^\nu\leq \bld \beta_\nu \}.\label{Newt_component}
\end{equation}
For $p(\bld x) = \sum_{\nu=1}^N p^\nu(\bld x)$, the Newton polytope is given by 
$$\text{Newt}(p)= \{\bld F^1\bld \alpha^1+\dots+\bld F^N\bld \alpha^N: \bld {A}^\nu \bld \alpha^\nu\leq \bld \beta_\nu,\nu=1,\dots,N\},$$
which corresponds to the Minkowski sum $\text{Newt}(p^1)\oplus\dots\oplus\text{Newt}(p^N). $
For $p(\bld x) = \bigvee_{\nu=1}^N p^\nu(\bld x)$, the Newton polytope is given by \begin{align*}
\text{Newt}(p)=& \{\bld F^1\bld \alpha^1+\dots+\bld F^N\bld \alpha^N: \exists \lambda^\nu\geq 0,\nu=1,\dots,N, \\&\bld {A}^\nu \bld \alpha^\nu\leq \bld \beta_\nu \lambda^\nu,~ \sum_{\nu=1}^N \lambda^\nu=1\}.
\end{align*}
 
 \subsection{Neural Networks As Differences of Composite Polynomials}

We then present some examples of neural networks represented as the difference of two composite tropical polynomials. 

\begin{example}[Single hidden layer ReLU network]
\label{DifferenceOfTropPolExample}
This example follows \cite{smyrnis2019tropical}. 
Consider a neural network with $n$ inputs, a single hidden layer with $m_p$ neurons and ReLU activation functions, and a single linear output neuron. 
The output of the $\nu$-th neuron of the hidden layer is:
$$z_\nu=\max(\bld w_\nu^T \bld x+\beta_\nu,0) $$
The output $y$ can be written as 
$$y = \sum_{\nu=1}^{N_1} w^{2+}_{\nu}z_\nu -\sum_{\nu=N_1+1}^{N} w^{2-}_{\nu}z_\nu+\beta^2, $$
where $N_1+N_2=N$, and  $N_1, N_2$ are the number of neurons that have positive weight and negative weights respectively. Without loss of generality, we assume that the neurons are ordered such that the weights of the first neurons to be positive and the weights of the last negative. The output of the neural network can be expressed as the difference of two tropical polynomials $p_1(\bld x)$, $p_2(\bld x)$, plus a constant term. Each of these tropical polynomials is written as the sum of simpler ones
\begin{align*}
p_1(\bld x)&= \sum_{\nu=1}^{N_1} \max(w^{2+}_{\nu}\bld w_\nu^T \bld x+w^{2+}_{\nu}\beta_\nu,0) \\&= \sum_{\nu=1}^{N_1} \max((\bld a_1^\nu)^T \bld x+b_\nu,0)\\
p_2(\bld x)&= \sum_{\nu=N_1+1}^{N} \max(w^{2-}_{\nu}\bld w_\nu^T \bld x+w^{2-}_{\nu}\beta_\nu,0) \\&= \sum_{\nu=N_1+1}^{N} \max((\bld a_2^\nu)^T \bld x+b_\nu,0)
\end{align*}
Note that $p_1$, $p_2$, expressed in their canonical form \eqref{Pdef}, can have a large number of terms, corresponding to linear regions of tropical polynomial functions $p_1$, $p_2$ (for an enumeration of the linear regions see \cite{zhang2018tropical,charisopoulos2018tropical}).

The Newton polytope of the simple tropical polynomial $\max((\bld a_1^\nu)^T\bld x+b_\nu,0)$ is the line segment $[0,\bld a_1^\nu]$ in the $d$-dimensional space. This polytope is equivalently described in the form \eqref{Newt_component} as $\text{Newt}(p^\nu) = \{ \alpha^\nu \bld a^\nu_1 : 0\leq \alpha^\nu\leq 1\} $. Thus, the Newton  polytope of $p_1$ is given by
$$\text{Newt}(p_1)= \{\alpha^1 \bld a^1_1+\dots +\bld  \alpha^1 \bld a^{n_1}_1: 0\leq \alpha^\nu\leq 1\},$$
which is a zonotope (see also \cite{charisopoulos2018tropical}). 

The constraint $\hat {\bld a}_i\in C$ of problem \eqref{Prob3} becomes
$$\hat{\bld a}_i +\tilde {\bld a}_j= \sum_{l=1}^n \bld a^{l} \lambda_{j,l}, ~~
0\leq \lambda_{j,l}\leq 1.$$

Finally, note that if the number of neurons of the hidden layer $N_1,N_2$ is smaller than the input dimension $n$, then the tropical polynomials $p_1,p_2 $ can be expressed in dimensions $N_1,N_2$.
\end{example}

\begin{example}[Multi-class to binary classification with a ReLU network]
\label{multiToBinaryExample}
Assume a neural network with a single ReLU hidden layer with $N$ units and a softmax output layer with $K$ units trained to classify samples to $K$ classes. The equations are the following
\begin{align*}
z^1_\nu = \max((\bld w^1_\nu)^Tx+b^1_\nu,0), ~~
z_k^2 =(\bld w_k^2)^T\bld z^1,~~
y_k = \frac{e^{z^2_k}}{\sum_{k'} e^{z^2_{k'}}}
\end{align*}
 We then assume, that for a given set of  samples, we know that the correct class is either $i_1$ or $i_2$. The neural network can be simply reduced for binary classification, substituting the last layer with a single neuron with sigmoid activation with output given by
 $$y =\sigma((\bld w_{i_1}^2-\bld w_{i_2}^2)^T\bld z^2+\beta^2),$$
 where $\sigma(\cdot)$ is the sigmoid activation function,, i.e. $\sigma(z)=1/(1-e^{-z})$.
 Note that this transformation corresponds to the application of Bayes' rule with prior probability of $1/2$ for classes $i_1,i_2$, and likelihoods given by the output of the initial neural network. 
 
 The output of the new network can be represented as a difference of two tropical polynomials as in the previous example. 
 
\end{example}

\begin{example}[Representing a multi-class network as a vector of tropical polynomials] 
\label{MultiClassExample}
Consider a neural network with a single-hidden layer having ReLU activations and a linear output layer with $K$ neurons, followed by softmax. Each output (before the softmax) $z^2_k$ can be expressed as a difference of two tropical polynomials plus a constant term 
$$
z^2_k=p_k^+(\bld x)-p_k^-(\bld x)+b_k.$$
The output is given by
$$
y_k = \frac{e^{z^2_k}}{\sum_{k'} e^{z^2_{k'}}}.
$$
Let us add to each of $z^2_k$'s the sum of all negative polynomials. We get the tropical  polynomials
\begin{equation}\tilde z^2_k=p_k^+(\bld x)+\sum_{k'\neq k}p_{k'}^-(\bld x)+b_k.\label{Omonyma}
\end{equation}
Then, it is easy to see that 
$$y_k = \frac{e^{\tilde z^2_k}}{\sum_{k'} e^{\tilde z^2_{k'}}}.$$
Finally, denoting by $z_i^1$, $i=1,...M$  the output of the hidden layer before the ReLU, then $\tilde z_k^2$ can be written in the form
\begin{equation}
\tilde z_k^2=\sum_{i=1}^M\bar w_{k,i}\max(z_i^1,0),
\label{sum_form}
\end{equation}
for appropriate non-negative constants $\bar w_{k,i}$. 
\end{example}

\begin{remark}
In the tropical framework, the difference of two polynomials  is considered as a tropical rational function. The addition of all the negative polynomials (denominators) corresponds to making the tropical fractions have the same denominator. 
\end{remark}

\subsection{ Division Algorithm for Composite Quotient} 
\label{Sec:CompositeQuotient}
We then describe an algorithm to divide a composite tropical polynomial $p(\bld x)=\sum_{i=1}^n\max(\bld a^T_i\bld x+b_i,0)$, where $n$ is the dimension
of the underlying space, by the zero polynomial $d(\bld x)=0$ and search for a quotient in the form $q(\bld x)=\sum_{i=1}^{\tilde m_{q}}\max(\hat{\bld a}^T_i\bld x+\hat b_i,0)$.  
Such polynomials occur in single hidden layer ReLU neural networks if the
number of neurons in the hidden layer is smaller than the space
dimension, using  a dimensionality reduction transformation (e.g., QR).

We assume that the vectors $\bld a_i$ are linearly independent. Denote by $\bld A$ the square matrix  of vectors $\bld a_i$, i.e., $\bld A=[\bld a_1~\dots\bld a_n]$, and by $\bld B$ the vector of $b_i$, i.e., $\bld B=[b_1~\dots~b_n]^T$. 

The algorithm is based on the fact that the quotient $q$ is the largest tropical polynomial function which is less than or equal to $p$. Consider a set of sample points $x^1,\dots,x^N$. Then the problem of maximizing $q$ can be approximated by
\begin{align}
\begin{aligned}
& \underset{ \hat {\bld a}_i,\hat b_i, i=1,\dots,m_{\tilde m_q}}{\text{maximize}}
& & \sum_{j=1}^{N}\sum_{i=1}^{m_{q}} \max(\hat {\bld a}_i^T\bld x^j+\hat b_i,0)\\
& \text{subject to}
& & q(\bld x)\leq p(\bld x), ~~\text{for all } x\in \mathbb R^n
\end{aligned}~~,
\label{Prob_Comp_quotient}
\end{align}

The following proposition reformulates the constraint $q(\bld x)\leq p(\bld x), ~~\text{for all } x\in \mathbb R^n$ in terms of the variables $\hat {\bld a}_i,\hat b_i, i=1,\dots,m_{q}$. 

\begin{proposition}
\label{Ineq_Comp_Quotient_Prop}
Let $p(\bld x), q(\bld x) $ as above. Then, $q(\bld x)\leq p(\bld x)$, for all $x\in \mathbb R^n$ if and only if
\begin{align}
\bld 0 \leq \bld A^{-1} \hat{\bld a}_i,~~
\bld A^{-1} \sum_{i=1}^{m_{\tilde m_q}} \hat{\bld a}_i \leq \bld 1,~~
\hat{b}_i\leq \bld B^T \bld A^{-1} \hat{\bld a}_i
\label{PropIneqConditionsCompQuot}
\end{align}
 for $i=1,\dots,m_q$, where $\bld 0$ is the vector having all its entries equal to zero and $\bld 1$ all entries equal to  one.
\end{proposition}

\textit{Phase 1}: For the set of sample points $\bld x^1,\dots, \bld x^N$ we compute $y_{i,j} = \chi_{\hat {\bld a}_i\bld x^j+b_i\geq 0}$, where $\chi$ is the indicator function (i.e., $\chi_{\hat {\bld a}_i\bld x^j+b_i\geq 0}=1$ if $\hat {\bld a}_i\bld x^j+b_i\geq 0$ and $0$ otherwise). Based on the values of $y_{i,j}$, we determine the local form of the objective in \eqref{Prob_Comp_quotient}. Then, we compute $\bld c_1$ and $\bld c_2$ as $$\big[\sum_{j=1}^N y_{1,j}\bld x_j^T~\dots~\sum_{j=1}^N y_{m_{\tilde m_q},j}\bld x_j^T\big],~\big[\sum_{j=1}^N y_{1,j}~\dots~\sum_{j=1}^N y_{m_{\tilde m_q},j}\big].$$  

\textit{Phase 2}: We solve the linear programming problem  
\begin{align}
\begin{aligned}
& \underset{ \hat {\bld a}_i,b_i,i=1,\dots,m_{\tilde m_q}}{\text{maximize}}
& & \bld c_1 \left[  \begin{matrix}\hat{\bld a}_1\\\vdots\\ \hat{\bld a}_{m_{\tilde m_q}} \end{matrix} \right]+\bld c_2 \left[  \begin{matrix}\hat{ b}_1\\\vdots\\ \hat{ b}_{m_{\tilde m_q}} \end{matrix} \right]\\
& \text{subject to}
& &  
\bld 0 \leq \bld A^{-1} \hat{\bld a}_i, ~~\hat{b}_i\leq \bld B^T \bld A^{-1} \hat{\bld a}_i\\
& & &\bld A^{-1} \sum_{i=1}^{m_q} \hat{\bld a}_i \leq \bld 1,
\end{aligned}~~
\label{Prob5}
\end{align}
and denote its solution by $ \hat {\bld a}_i^{\text{new}},b_i^{\text{new}}$. We set 
\begin{align}
\begin{aligned} 
\hat {\bld a}_i &\leftarrow (1-\rho)\hat {\bld a}_i+\hat {\bld a}_i^{\text{new}}\rho, \\ 
\hat b_i&\leftarrow (1-\rho)b_i^{\text{}}+\rho b_i^{\text{new}},  
\end{aligned}\label{ALG_eq}
\end{align}
where $\rho\in(0,1)$ and go to Phase 1.
\begin{remark}
The algorithm corresponds to the application of  Frank–Wolfe (conditional gradient) method (see, e.g., \cite{bertsekas2016nonlinear}) to \eqref{Prob_Comp_quotient}. 
\end{remark}
 
Let us note that this linear programming problem \eqref{Prob5} is analytically solvable. Indeed, observing that the coefficients of $\hat b_i$'s are nonnegative, we can use the second inequality to compute the the optimal value of $\hat b_i$'s. Substituting back to the problem, we end up with $n$ independent linear programs. The feasible set of each of these problems has $\tilde m_q+1$ vertices. We obtain the optimal solution of the problem comparing the value of the linear objective on these vertices.

\begin{remark}
Let us comment on the difference of the algorithm of this subsection with the algorithm of Subsection \ref{Sec:Composite}. The algorithm of this subsection provides a solution a composite tropical polynomial, which corresponds to a (compressed) neural network with ReLu activations, while the algorithm of Subsection \ref{Sec:Composite} provides maxout solutions. 
\end{remark}

\subsection{Dividing a Vector of Composite Polynomials}
\label{SimplifiedVectorComposite}
In this subsection, we propose a simplified algorithm for dividing a vector of tropical polynomials by the zero polynomial. The motivation comes from Example \ref{MultiClassExample}. Consider a set of tropical polynomials in the form \eqref{sum_form}. 

We then describe a simplified version of the the linear optimization part of Algorithm \ref{alg:ApproxPolyDivi}. Consider a set of sample points $\bld z^1(1),\dots,\bld z^1(N) $. Observe that, with this formulation, both constraints \eqref{Prob3} are satisfied by any $\hat {\bld a}_{k}$, such that its $i-$th component $\hat a_{i,k}$  satisfies 
$0\leq\hat a_{i,k}\leq  \bar w_{i,k}$ and $\hat b_k=0$.  Then, the solution of  \eqref{Prob3}, for the $l-th$ term of the division of the $k-$th polynomial is approximated by the optimal solution of
\begin{align}
\begin{aligned}
& \underset{ \hat {\bld a}_{k}^l}{\text{maximize}}
& &  \bld s_{l,k}^T\hat {\bld a}_{k}^l\\
& \text{subject to}&&
0\leq\hat a_{i,k}^l\leq  \bar w_{i,k},~~ i=1,\dots,M
\end{aligned}~.
\label{ParallelogramConstr}
\end{align}
The solution of \eqref{ParallelogramConstr} is given by:
$$\hat a_{i,k}^l=\begin{cases}\bar w_{i,k}  \text{~~~ if ~}   s_{i,l,k}>0\\
0\text{~~~~~~ otherwise}\end{cases}$$

\section{Numerical Results}
\label{Sec:Numerical}

This section, presents some numerical examples for the tropical division algorithms. First, we present some simple examples to build some intuition for the behaviour of Algorithm \ref{alg:ApproxPolyDivi}. Then,  some applications of tropical division to the compression of neural networks are presented.  We focus on MNIST handwritten and CIFAR-10 datasets. 

\subsection{Numerical Examples for Algorithm \ref{alg:ApproxPolyDivi}}
As a first example, we present the application of  Algorithm \ref{alg:ApproxPolyDivi} to the tropical polynomial  division of Example \ref{ExactExample}. 
We choose $200$ sample points distributed according to the normal distribution with unit variance $\mathcal N(0,I_2)$. The algorithm converges in two steps and gives an approximate solution\footnote{The code for the numerical experiments is available online at https://github.com/jkordonis/TropicalML}.
\begin{align*}
\hat q(x,y ) = &\max(1.5038x +1.4962y+    0.0182,\\& 0.0003x+0.0288, 3x+  0.0241),
\end{align*}
which is very close to the actual quotient (see Example \ref{ExactExample}). 
Running again the algorithm with larger sample sizes we conclude that increasing the number of sample points reduces the error. 

\begin{figure}
\centering
\includegraphics[width=0.4\textwidth]{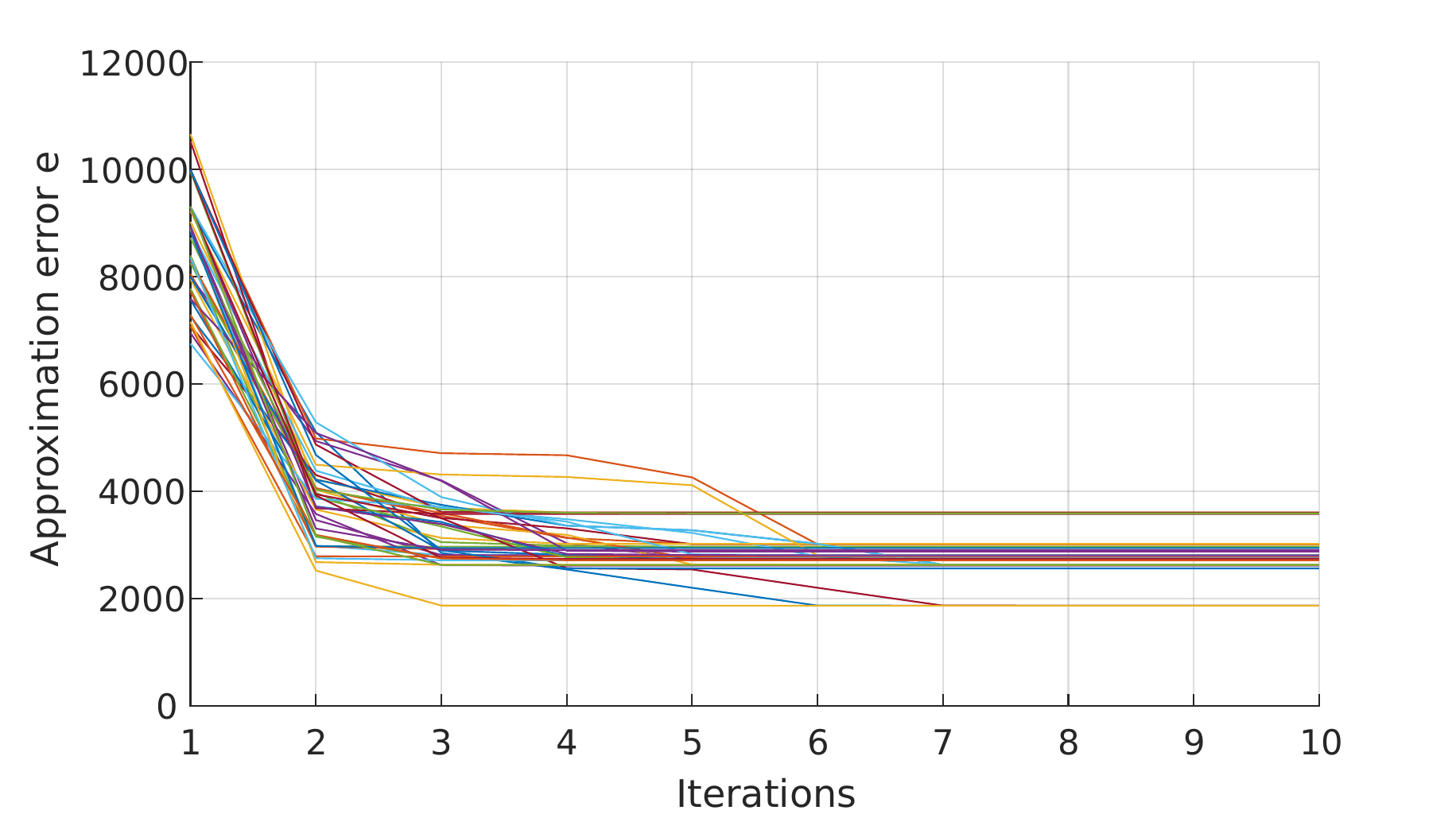}
\caption{\textit{The total error $e$ for executions of Algorithm \ref{alg:ApproxPolyDivi}, for different initial partitions.}}
\label{five_ponts}
\centering
\includegraphics[width=0.4\textwidth]{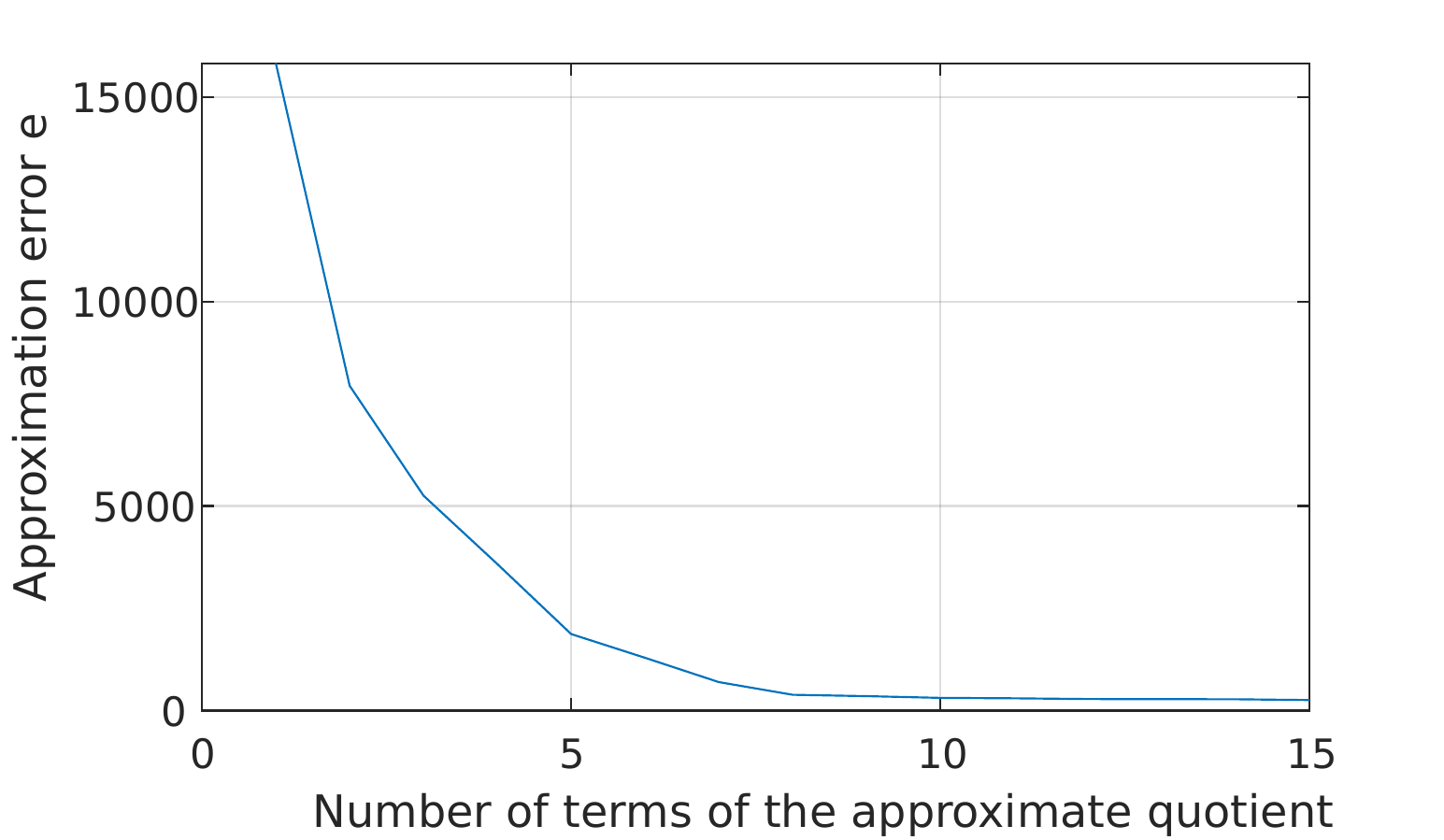}
\caption{\textit{The total error $e$ for divisions having different number of terms for the approximate quotient}}
\label{var_numb_of_terms}
\end{figure}

As a second example we divide a random tropical polynomial with $m_p=128$ terms in $n=3$   dimensions with another having $m_d=2$ terms. First, we run Algorithm \ref{alg:ApproxPolyDivi}, for $\tilde m_q=5$, with multiple initial partitions. The evolution of the error for the several runs is shown in Figure \ref{five_ponts}. The results illustrate the need for a multi-start method.  Indeed, for different random initial partitions the algorithm converges to different local minima. 
The quality of the approximation $e$ after 10 steps of the algorithm is given in Figure \ref{var_numb_of_terms} for a varying number of terms $\tilde m_q$. This result shows that we may derive a good approximation of the division using a small number of terms (e.g. $8$).

\subsection{ MNIST Dataset}
This example considers the MNIST handwritten digits data set. It consists of $60000$  training examples and $10000$ test examples of $28\times 28$ gray scale images, that represent digits $0-9$. 
We start with a single hidden layer  neural network with 100  hidden units  with ReLU activations and $10$ softmax output units. The network is trained in the original training data set using standard techniques (Adam optimizer on cross-entropy loss, batch size of 128). We then design a smaller network discriminating between digits $3$ and $5$. The input space has $28\cdot 28=784$ dimensions. 

We first use the technique described in Example \ref{multiToBinaryExample} to reduce the output layer of the original network to a single neuron. Then, using  the technique described in Example \ref{DifferenceOfTropPolExample}, we express the output (before the sigmoid) as the difference of two tropical polynomials. These polynomials are expressed as the summation of $100$ terms in the form $\max(\bld\alpha_l^\nu\bld x,0)$, where $l=1,2$. We then apply a reduced QR decomposition to the matrices $\bld A_1=[\bld\alpha_1^1~\dots,~\bld\alpha_1^{100}]$ and $\bld A_2=[\bld\alpha_2^1~\dots,~\bld\alpha_2^{100}]$, writing them as $\bld A_1=\bld Q_1\bld A_1^r$ and $\bld A_2=\bld Q_2\bld A_2^r$. The $\bld A_l^r$ matrices have as columns $\bld \alpha _l^{r,\nu}$, i.e., an expression of the vectors $\bld \alpha _l^{\nu}$ in the subspace of their span. For each polynomial the input has been transformed as $\bld x_l = \bld Q_l^T \bld x$ (recall Remark \ref{RedDimRem}). 

We divide each of these polynomials with the zero polynomial $d =0$, applying two different variants of the Algorithm 2. In both cases, we use as sample points the first $200$ training points. The first variant  (modification of Section \ref{Sec:Composite}), the approximate quotients have the form $q_l(\bld x_l) = \bigvee_{i=1}^{\tilde m_q} (\hat {\bld a}_i^l \bld x_l+\hat b_i^l)$, $l=1,2$.  Then, the output of the original network can be approximated as
$$y^{\text{appr}}=\sigma(q_1(\bld Q_1^T \bld x)-q_2(\bld Q_2^T \bld x)+\beta^2)$$ 

The computation of the approximate output corresponds to a neural network an input layer, a hidden layer with two maxout units, a linear layer with a single unit (computing the difference of the two maxout units) and an output layer with a single sigmoid unit. 

The second variant, uses approximates the quotients in the form $\sum_{i=1}^{m_{q}} \max(\hat {\bld a}_i^T\bld x+\hat b_i,0)$ and uses the modification of subsection \ref{Sec:CompositeQuotient}. The compressed network in this case is simply a smaller ReLU network.  

Table \ref{Table_MNIST} presents the error rate on the test set  for the original neural network, as well as the computed maxout and ReLU networks. We compare the reults when each maxout unit has $3$, $5$, and $10$ terms respectively and each of the tropical polynomials (positive and negative) has $3$, $5$, and $10$ ReLU units. It also presents the percentage of parameters that remain after the compression. As a baseline for comparison, we use structured L1 pruning without retraining (see e.g. \cite{blalock2020state}). That is, this method deletes the neurons having weights with small L1 norm. We compare reduced neural networks with the same number of parameters. We present two results, the binary comparison of digits $3$ and $5$ and an average of the $45=10\cdot 9/2$ different binary comparisons.
 
\begin{figure}
\centering
\includegraphics[width=0.45\textwidth]{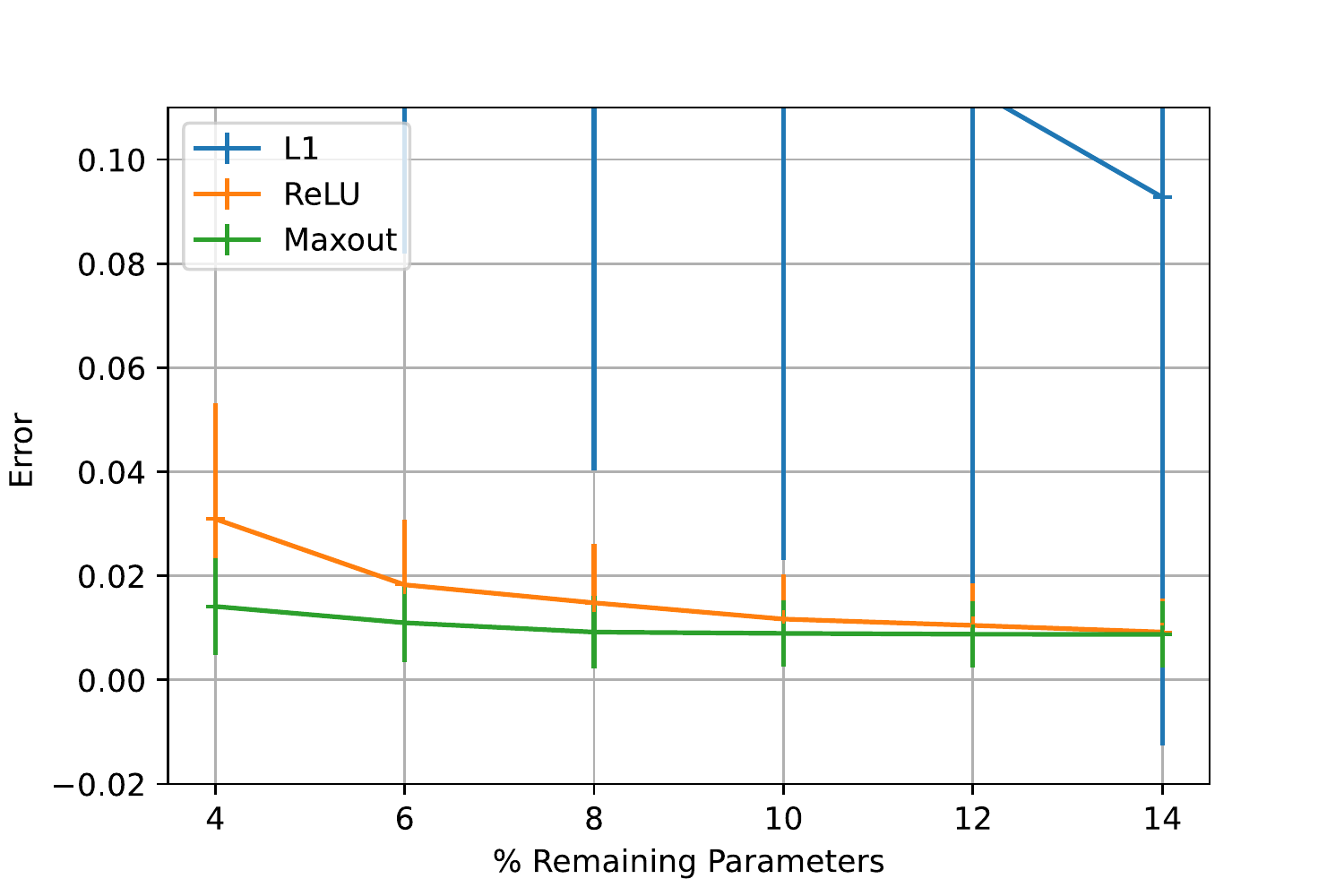}
\caption{\textit{ The average error rate of the compressed networks for the binary MNIST classification, compared with L1 structured pruning. Both techniques lead to better accuracies compared to L1 pruning. Furthermore, maxout networks clearly outperform both the L1 structured pruning and the ReLU compressed network, especially in the large compression regime. }}
\label{ReLUvsMaxoutvsL1}
\end{figure}

 The performance of the algorithms is shown in Figure \ref{ReLUvsMaxoutvsL1}. 
We observe that the algorithm of  Section \ref{Sec:Composite} produce superior results compared to the algorithm of Section \ref{Sec:CompositeQuotient}. Furthermore, both algorithms outperform L1 structured pruning, especially in the high compression regime. Let us note that the linear programming
problem of Section \ref{Sec:CompositeQuotient} is analytically solvable, and thus the corresponding algorithm is faster.

\mycomment{
\begin{table}[h!]
  \begin{center}\scalebox{0.8}{
    \begin{tabular}[h!]{ c | c |c| c| c| c} 
Network& Orig. 10 cl. & Orig. 2 cl. & $\tilde m_q=3$&  $\tilde m_q=5$& $\tilde m_q=10$\\  
\hline
Error Div.& 2.19\% & 1.21\% & 2.15\% &1.58\%&1.74\%
\\\hline
Error L1 &    2.19\% & 1.21\% &25.1\% &12.5\% & 2.05
\\\hline
\# Param. &  77910& 77001& 4615& 7691& 15381 \\ \hline
 \begin{tabular}{@{}c@{}}\% of Param.\\ Remaining \end{tabular} & 100\% &100\%&6\%&10\%&20\%
\end{tabular}
 }
    \caption{\textit{Results for the 3-5 pair of the MNIST dataset. Error rate of the original network and the reduced  networks. Error Div. represents the error rate for the reduced networks obtained using the tropical division algorithm and Error L1 the error rate of with the L1 structured pruning algorithm. }}
    \label{Table_MNIST}
  \end{center}
\end{table}}

\begin{table}[h!]
  \begin{center}\scalebox{0.8}{
    \begin{tabular}[h!]{ c  |c| c| c| c} 
Network. & Orig. 2 cl. & $\tilde m_q=7$                        &   $\tilde m_q=5$& $\tilde m_q=3$\\
\hline
Err. L1 Avg. &  0.4$\pm$ 0.25\% & 9.27$\pm$10.53\%   &  13.76 $\pm$ 11.46\%&21.37$\pm$ 13.84\%
\\  
\hline
Err. Maxout. Avg. & 0.4$\pm$ 0.25\% & \textbf{0.87$\pm$0.63}\%  &\textbf{0.89$\pm$ 0.63}\%& \textbf{1.1$\pm$0.75}\%
\\  
\hline
Err. ReLU. Avg. & 0.4$\pm$ 0.25\% & {0.92$\pm$0.65}\%  &{1.17$\pm$ 0.85}\%& {1.25$\pm$1.26}\%
\\\hline
Err. L1 3-5  & 0.79\% &4.83\%&16.35\% & 17.98\% 
\\\hline
Err. Maxout. 3-5 &1.05\% & \textbf{2.05}\%  &\textbf{1.84}\%&\textbf{2.21}\%
\\\hline
Err. ReLU. 3-5 &1.05\% & \textbf{2.05}\%  &{2.68}\%&{3.83}\%
\\\hline
\# Param.& 77001&15381  & 7691& 4615\\ \hline
 \begin{tabular}{@{}c@{}}\% of Param.\\ Remaining \end{tabular} & 100\%&20\%&10\%&6\%
\end{tabular}
 }
    \caption{\textit{Results for the MNIST dataset. Error rate of the original network and the reduced  networks. Err. Maxout represents the error rate for the reduced networks obtained using the tropical division algorithm of  Section \ref{Sec:Composite}, Err. ReLU.  the error of the network obtained using the tropical division algorithm of  Section \ref{Sec:CompositeQuotient}, and Err. L1 the error rate of with the L1 structured pruning algorithm. The error rates for the 3-5 binary comparison and the average over the $45$ binary comparisons are presented.}}
    \label{Table_MNIST}
  \end{center}
\end{table}

 \begin{remark}
 It is worth noting substituting $\bld x_l$ into $q_l$, we get $$q_1(\bld Q_1^T \bld x) = \bigvee_{i=1}^{\tilde m_q} (\hat {\bld a}_i^l Q_l^T \bld x+\hat b_i^l).$$
Thus we do not need to store $Q_1,Q_2$ but only the vectors $\hat {\bld a}_i^1 Q_l^T$, and the scalars $\hat b_i^l$,  $\beta^2$, for $l=1,2$ and $i=1,\dots,m_q $.
\end{remark}

\begin{remark}
 Let us note that for computing the reduced networks, we used only the first 200 samples of the training data. 
 \end{remark}

In the following, we continue exploring the  algorithm of  Section \ref{Sec:Composite} to more complex examples.

\mycomment{

\subsubsection{Compression of a ReLU Network}
In this section, we approximate the MNIST ReLU  network of the previous section with a smaller ReLU network. We divide each polynomial $p_1$ and $p_2$ by the zero polynomial and search for quotients in the form  $q(\bld x)=\sum_{i=1}^{\tilde m_q}\max(\hat{\bld a}^T_i\bld x+\hat b_i,0)$. 
We use the algorithm described in Subsection \ref{CompQuotSection}, using as sample points the first 200 data points of the training set. The results are reported in Table \ref{Table_MNIST_ReLU2ReLU}.

\begin{table}[h!]
  \begin{center}\scalebox{0.9}{
    \begin{tabular}[h!]{ c|  c |c| c| c|c} 
Network& Orig. 10 cl. & Orig. 2 cl.&   $\tilde m_q=3$&  $\tilde m_q=5$&  $\tilde m_q=10$\\ 
\hline
Error rate& 2.2\% & 1\% &2.67\% &2.1\% & 1.94\%
\end{tabular} }
    \caption{\textit{Error rate of the original network and the reduced ReLU network.}}
    \label{Table_MNIST_ReLU2ReLU}
  \end{center}
\end{table}

Comparing tables \ref{Table_MNIST} and \ref{Table_MNIST_ReLU2ReLU}, we conclude that the  ReLU network in the MNIST digit dataset is more accurately approximated using maxout units compared to ReLU units. 
}
\subsection{CIFAR-10 Dataset-Binary}
This example concerns CIFAR-10 dataset which contains small ($32\times 32$) color images. The training set consists of 50000 images of 10 classes (5000 samples for each class) and the test set of 10000 images (1000 samples per class). We first train a VGG-like network \cite{simonyan2014very}, having three blocks consisting of two convolution layers with ReLU activation followed by a $2\times 2$ max-pooling layer with padding, a dense layer with 1024 neurons, and an output layer with 10 neurons. The convolution layers in the first, second, and third block have 32, 64, and 128 neurons, respectively. The network was trained using standard techniques (Adam optimizer for the cross-entropy loss, batch normalization and dropout for regularization and data augmentation). Let us note that the input of the dense layer has dimension $4\cdot4\cdot128=2048$.

We then design a simplified neural network discriminating between pairs of classes (e.g., ``Automobile'' and ``Truck'').  To do so, we approximate the dense layer with 1024 neurons using the techniques of the previous section. Particularly, using the technique of Example \ref{multiToBinaryExample} we describe the output of the neural network as the difference of two tropical polynomials in $2048$ dimensions. These polynomials are then divided by the zero polynomial applying the Algorithm 2, with the modification of Section \ref{Sec:Composite}, and using as sample points the first $500$ training points.

Table \ref{Table_CIRAR10_ReLU2MaxOut} presents the error rate on the test set  when the maxout units on the dense layer have $3$, and $5$ terms respectively. Three tests results are presented: on pairs Automobile-Truck (A-T), Cat-Dog (C-D) and the average of the 45=$9\cdot10/2$ possible binary comparisons. The results obtained with the tropical division algorithm are compared with the L1 structured pruning. Table \ref{Table_CIRAR10_ReLU2MaxOut}  also presents the number of parameters of the dense layer of the networks. We observe that the dense layer can be compressed to less than  $1\%$ of its original size with minimum performance loss. 
\begin{table}[h!]
  \begin{center}\scalebox{0.9}{
    \begin{tabular}[h!]{ c||  c| c| c} 
Network  & Orig. 2 classes &  $\tilde m_q=5$&  $\tilde m_q=3$ \\ 
\hline
Err. L1 A-T &4.04\% &25.8\%   &46.45\%
\\ \hline
Err. Div. A-T  & 4.05\% & \textbf{4.6}\% & \textbf{4.4}\% 
\\\hline
Err. L1 C-D  &13.7\% &49.3\%  & 50\%
\\\hline
Err. Div. C-D & 13.7\% &\textbf{15.4}\%  & \textbf{16.85}\%
\\\hline
Err. L1 Avg.  & 2.74$\pm$2.57\% &43.45$\pm$10.57\% &43.62$\pm$10.61\%  
\\\hline
Err. Div. Avg. & 2.74$\pm$2.57\% &  \textbf{3.85$\pm$3.16}\%  & \textbf{4.06$\pm$3.37}\%
\\\hline
\# Param. &$2.1\cdot10^6$& $2\cdot10^4$   & $1.2\cdot10^4$  \\\hline
 \begin{tabular}{@{}c@{}}\% of Param.\\ Remaining \end{tabular}  &100\%&0.95\%&0.57\%
\end{tabular} }
    \caption{\textit{Results for the CIFAR-10 dataset. Error rate of the original network, the reduced maxout networks and the reduced networks obtained using the L1 structured pruning algorithm. We present the Automobile vs Truck case, the Cat vs Dog case, and the average of the $45=10\cdot9/2 $ different binary comparisons. The results are indexed using A-T, C-D, and Avg. }}
    \label{Table_CIRAR10_ReLU2MaxOut}
  \end{center}
\end{table}

\begin{remark}
The proposed scheme is very competitive for the large compression regime, that is, in the case where there are very few parameters remaining. Furthermore, it can be performed assuming access to a small portion of the training data (500 out of 50000).
\end{remark}

\subsection{CIFAR-10 Multiclass}

We then compress the multi-class neural network trained in the previous subsection for the CIFAR-10 dataset. 
As the input to the compression algorithm, we consider the output of the convolutional part of the neural network with dimension 2048.

We first represent the logits (the output before the softmax) as a vector of tropical polynomials using \eqref{Omonyma} in 
Example \ref{MultiClassExample}. We then simplify each of the tropical polynomials using the simplified algorithm of Section \ref{SimplifiedVectorComposite} and a random sample of $500$ training points. The results are summarized in Table \ref{TableCIFAR10Multi}. Additionally, we present an improved prediction scheme, where the $K\times \tilde m_q$ values $(\bld {\hat a}_k^l)^T\bld z^1$, with $k=1,\dots,K$ and $l=1,\dots,\tilde m_q$ are fed to a small single hidden layer ReLU neural network with 100 units.  We refer to this small neural network as head. 

 \begin{table}[h!]
  \begin{center}\scalebox{0.9}{
    \begin{tabular}[h!]{ c|  c |c| c| c} 
Network& Original& $\hat m_q=7$&  $\hat m_q=5$&  $\hat m_q=3$\\ 
\hline
Error L1 &   14.18\%&45\% & 57.3\%&71.7\%  \\\hline
Error Div.& 14.18\%&   41$\pm 1.8$ \%         & 40.1 $\pm$ 1.7\%& 42.5  $\pm$ 2\%\\\hline
Err. Div. Improved& 14.18\%&\textbf{26.6$\pm$ 0.3 }\%            & \textbf{26.4 $\pm$ 0.3}\%& \textbf{28.1  $\pm$ 0.1}\%\\\hline
\# Param. &$2.1\cdot10^6$    & $1.4\cdot10^5 $  &  $10^5 $  & $ 6.6\cdot10^4$   \\ \hline
 \begin{tabular}{@{}c@{}}\% of Param.\\ Remaining \end{tabular}  & 100\% &  7.32\%& 5.24\% &3.16\%
\end{tabular}   
 }
    \caption{\textit{Results for the compression on the CIFAR-10 dataset. Average error rates for multiple executions. Comparison of the original division algorithm, the improved division algorithm and the L1 pruning. The number of parameters include also the parameters of the head. }} \label{TableCIFAR10Multi}
  \end{center}
\end{table}
Note that the additional parameters for the improved prediction are very few. Indeed for the case of $\hat m_q=3,5,7$ we have $4110$, $6110
$ and $8110$ parameters respectively. 

We observe again that the division algorithm works well on the large compression regime. Furthermore, in this example there is no improvement when we use more terms in the quotient polynomials. This is probably due to stacking in local optima. 

\begin{figure}
\centering
\includegraphics[width=0.45\textwidth]{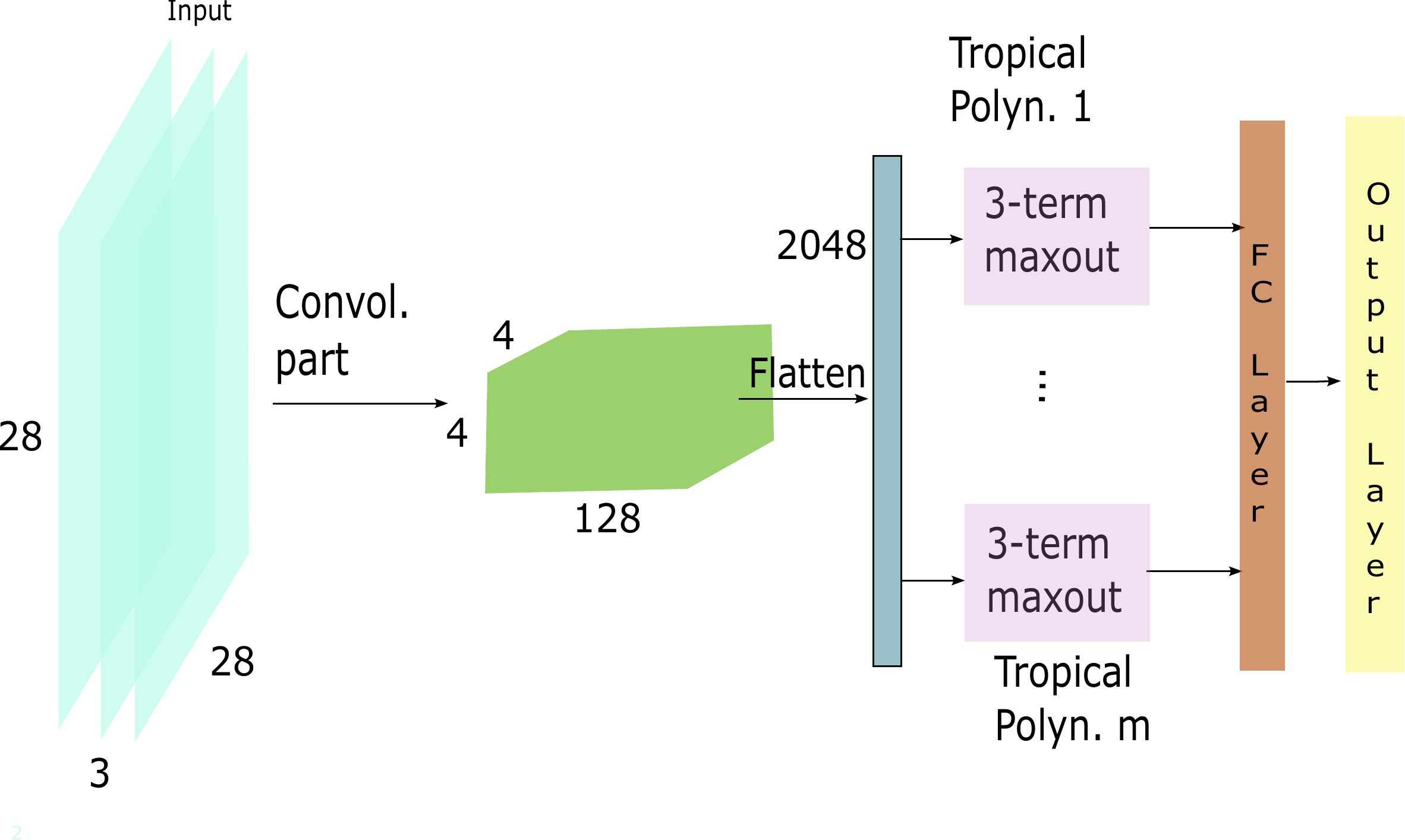}
\caption{\textit{The structure of the proposed reduced network. Each pink box consists of linear neurons obtained by successive implementations of the division algorithm.}}
\label{Reduced_NN}
\end{figure}

We then exploit the good behaviour of the binary classification algorithm and apply it to the multi-class problem. To do so, we use a subset of the tropical polynomials obtained for binary classification.  In each binary comparison, the division algorithm gives two tropical polynomials and the $9\times10/2=45$ binary comparisons give a total of $90$ tropical polynomials. The value of a subset of the tropical polynomials is fed to a single hidden layer network with ReLU activations, as described in Figure \ref{Reduced_NN}. 

\begin{figure}
\centering
\includegraphics[width=0.45\textwidth]{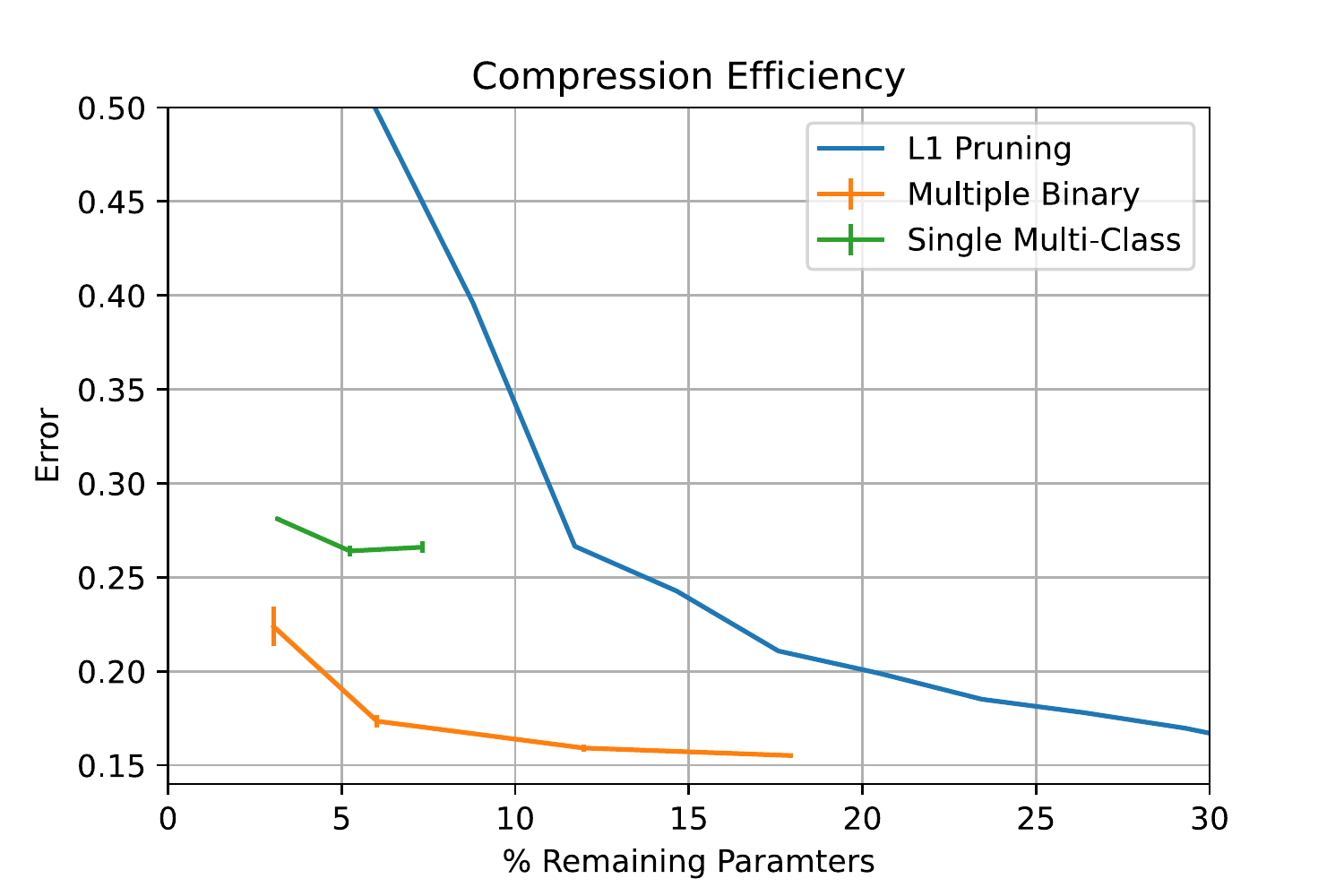}
\caption{\textit{The error rate of the compressed neural networks obtained using the multiple division algorithm. The blue and yellow lines use divisions with $\tilde m_q=3$ and $\tilde m_q=5$, respectively. The horizontal axis represents the percentage of the remaining parameters. Finally, the green line corresponds to using random vectors instead of the quotients.}}
\label{ErrorLines}
\end{figure}

Figure \ref{ErrorLines} presents the error of the compressed networks obtained using the multiple binary division algorithm of the previous paragraph. To produce these results we used a random set of $m$ tropical polynomials with $m=10,20,40,60$.  

We observe that the proposed method works better than L1-structured pruning, when the number of remaining parameters is small.

\begin{remark}
Note that in all the examples, the tropical division compression algorithms used only a small subset of the training data. This can be useful when simplifying a network where only parts of the training dataset are available. 
\end{remark}

\begin{remark}
An interesting point of future research is to investigate whether producing additional sample points with the usual data augmentation techniques improves the final result. 
\end{remark}

\section{Conclusion}
This work proposed a new framework for tropical polynomial division and its application to neural networks. We showed the existence of a unique quotient-remainder pair and characterized the quotient as the convex bi-conjugate of the difference of the dividend from the divisor. This characterization led to an exact algorithm based on polyhedral geometry. We also proposed an approximate algorithm based on the alternation between clustering and linear programming. We then focused on dividing composite tropical polynomials and proposed two modified algorithms, one producing simple quotients and the other composite. Finally, we applied tropical polynomial division techniques to simplify  neural networks with ReLU activations. The resulting neural networks have either maxout or ReLU activations, depending on the division algorithm chosen. The results are promising and compare favorably with a simple baseline (L1 pruning).

There are several directions for future research. First, we may combine L1 regularization to the problem \eqref{Prob4} to induce sparse solutions and improve further neural network compression. Another direction is to study compression problems for neural networks with many outputs, using a single division. A possible tool in this direction is the Cayley trick \cite{sturmfels1994newton,joswig2017cayley}. 
Finally, the study of alternative optimization algorithms is of certain interest.

\section*{Acknowledgment} \addcontentsline{toc}{section}{Acknowledgment}

The authors are grateful to Dr. George Retsinas for his valuable suggestions and comments.

%\mycomment
{

\appendix
\section{\appendixname\,: Proof Of Propositions of the Main Text}

\subsection{Proof of Proposition \ref{LemmaBackground}}
\label{monotonic_Prop}
 
The proof follows closely  the proof of Proposition 3.1.6 in \cite{maclagan2021introduction}.
Observe that
\begin{align}
\label{setEqA}
p_i(x) &= \max_{\bld a\in \text{Newt}(p_i)} \{  [\bld x^T ~1] [\bld a^T  ~   \text{ENF}_{p_i}(\bld a)        ]^T          \}\nonumber\\&= \max_{\bld a\in \mathbb R^n} \{  [\bld x^T ~1] [\bld a^T  ~   \text{ENF}_{p_i}(\bld a)        ]^T \}.
\end{align}

\mycomment{
The first equality of \eqref{setEqA} holds because
\begin{align*}
p_i(x)&=\max_j(\bld a_j^Tx+b_j) =\max_j([\bld a_j^T~b_j] [\bld x^T 1]^T) \\&= \max_{[\bld a_j^T~b_j]\in \text{ENewt}(p_i)} ([\bld a_i^T~b_i] [\bld x^T~ 1]^T)\\&=\max_{\bld a\in \text{Newt}(p_i)} \{  [\bld x^T ~1] [\bld a^T  ~   \text{ENF}_{p_i}(\bld a)        ]^T          \}. 
\end{align*}
The third equality of the last relation holds due to the fact that $[\bld a_i^T~b_i] [\bld x^T~ 1]^T$ is a linear function of $[\bld a_i^T~b_i] $ and $ \text{ENewt}(p_i)$ is a polytope, and the last from the definition of $\text{ENF}(p_i)$.
The last equality of \eqref{setEqA} holds because $\bld a\not\in \text{Newt} (p_i)$ implies   $ \text{ENF}_{p_i}(\bld a)=-\infty.$ 
}
Which implies that if $\text{ENF}_{p_1}(\bld a)\leq\text{ENF}_{p_2}(\bld a) $ then $p_1(\bld x)\leq p_2(\bld x),$ for all $\bld x\in \mathbb R^n.$
On the other hand, assume that $p_1(\bld x)\leq p_2(\bld x)$, and that for some $\bld a_0$ it holds $\text{ENF}_{p_1}(\bld a_0)>\text{ENF}_{p_2}(\bld a_0) .$ Then, the point  $[\bld a_0^T~ \text{ENF}_{p_1}(\bld a_0)]$ does not belong to the convex set $ C=\{[\bld a^T~ z]\in\mathbb R^{n+1}:z\leq  \text{ENF}_{p_2}(\bld a) \}.$
Thus, there is a $[\bld c_x^T ~c_z]\in\mathbb R^{n+1}$ such that 
$\bld c_x^T \bld a_0+ c_z\text{ENF}_{p_1}(\bld a_0)>\bld c_x^T \bld a+ c_z z,$
for all $[\bld a^T~ z] \in C$. Since $C$ is unbounded below, $c_z>0$. Furthermore, since $[\bld a^T~ \text{ENF}_{p_2}(\bld a)] \in C $ it holds
$$\bld a_0^T\bld x_0 + \text{ENF}_{p_1}(\bld a_0)>\bld a^T\bld x_0 +   \text{ENF}_{p_2}(\bld a),$$
for all $\bld a\in \mathbb R^n$, where $\bld x_0 = \bld c_x/c_z$.
Taking the maximum with respect to $\bld a$ and using \eqref{setEqA} in both sides we have:
\begin{align*}
p_1(\bld x_0)\geq \bld a_0^T\bld x_0 + \text{ENF}_{p_1}(\bld a_0)>&\max_{\bld a\in\mathbb R^n}\{\bld a^T\bld x_0 +   \text{ENF}_{p_2}(\bld a)\} \\&=p_2(\bld x_0),
\end{align*}
which is a contradiction.

\subsection{Proof of Proposition \ref{ExistProp}}
\label{Proof1}
It is not difficult to see that if $\tilde q (\bld x)= \bigvee_{i=1}^{m_{\tilde q}} (\tilde {\bld a}_i^T \bld x+\tilde b_i)$, and $\tilde q$ satisfies \eqref{QuotDef}, then $\tilde q (\bld x)\leq q (\bld x)$, for all $\bld x \in \mathbb R^n$. Indeed, since $$\tilde  {\bld a}_i^T \bld x+\tilde b_i\leq \tilde q(\bld x) \leq p(\bld x)-d(\bld x),$$ it holds $\tilde b_i\leq l(\tilde {\bld a}_i)$. Thus,
$$\tilde {\bld a}_i^T \bld x+\tilde b_i\leq \tilde {\bld a}_i^T \bld x+l({\bld a}_i)\leq q(\bld x).$$
Hence, $\tilde q(\bld x)\leq q(\bld x)$, for all $\bld x \in \mathbb R^n$. 

To prove that $q$ is a quotient, it remains to show that it is a tropical polynomial. The epigraph of $q$ is the closed convex hull of the epigraph of $f$ \cite{rockafellar2015convex}
\begin{align}
\text{epi}(q) = \overline{\text{conv}(\text{epi}(f))}.
\label{epigraph_ch}
\end{align}
 On the other hand, $f$ is a piecewise linear function. Thus, its epigraph is a union of a finite number  of convex polyhedra. Indeed if $A_1,\dots,A_{m_{f}}$ its linear regions (i.e., intersections of linear regions of $p$ and $d$) then $A_i,$ $i=1,...,m_f$ are convex polyhedra and 
$$\text{epi}(f) = \bigcup_{i=1}^{m_f} \{(\bld x,t)\in \mathbb R^{n+1}: \bld x\in A_i, t\geq f(\bld x)\}. $$
Observe that $\text{epi}(q)$ is a convex polyhedron and $q$ is a piecewise linear convex function. Thus, $q$ is a tropical polynomial. 
From the definition it is obvious that the quotient is unique.

Let $q$ be the quotient and assume that for some tropical polynomial $\tilde r$ it holds
$$p(\bld x)= (q(\bld x)+d(\bld x))\vee \tilde r(\bld x), \text{~~~~~~for all } \bld x\in \mathbb R^n.$$
Assume that for some point 
$\bld x_0$ it holds $p(\bld x_0)>d(\bld x_0)+q(\bld x_0)$ and that $p$ has the linear form $\bld a_{i_0}^T\bld x+b_{i_0} $ in a neighborhood of $\bld x_0$.
 Then, $\tilde r(\bld x)=p(\bld x)$ in this neighborhood. Furthermore, since $\tilde r$ is convex, we have
 $$\tilde r(\bld x)\geq \bigvee_{i\in I} \bld a_i \bld x+b_i=r(\bld x),$$
where $I = \{i\in \{1,\dots,m_p\}:\exists \bld x\in \mathbb R^n\text{ with } p(\bld x)>d(\bld x)+q(\bld x), \text{ and } p(\bld x) = \bld a_i \bld x+b_i \}$. It is not difficult to see that $r$ is a remainder.

\subsection{Proof of Proposition \ref{EffectiveProp}} 
\label{Proof2}
a)
Let $\bar q$ and $\bar r$ be the quotient and the remainder of the division of $r$ by $d$. Then
\begin{align}
r(\bld x)=(d(\bld x)+\bar q(\bld x))\vee \bar r (\bld x)
\label{doubleRem}
\end{align}
Combining with 
$$p(\bld x)=(d(\bld x)+q(\bld x))\vee r (\bld x),$$
we get
\begin{align}
p(\bld x)&=(d(\bld x)+q(\bld x))\vee(d(\bld x)+\bar q(\bld x))\vee \bar r (\bld x)\nonumber\\
&p(\bld x)=(d(\bld x)+q(\bld x)\vee\bar q(\bld x))\vee \bar r (\bld x)\label{COMBdiv}.
\end{align}
Equation \eqref{doubleRem} implies that $\bar r(\bld x)\leq r(\bld x)$. This fact combined with 
the fact that $q(\bld x)$ is the quotient of the division of $p(\bld x)$ by $d(\bld x)$ and \eqref{COMBdiv} implies that $\bar q(\bld x)\vee q(\bld x) = q(\bld x)$. Hence, since $r(\bld x)$ is the remainder of the division of $p(\bld x)$ by $d(\bld x)$ it holds $r(\bld x)\leq \bar r(\bld x)$. Hence, $r(\bld x)=\bar r(\bld x)$, for all $\bld x\in\mathbb R^n$ \hfill 

b) The proof is trivial.

\subsection{Proof of Proposition \ref{Prop_setC}}
\label{Proof3}
 Assume $p(\bld x)$ is given by \eqref{Pdef} and $d(\bld x)$ by
$$d(\bld x) = \bigvee_{i=1}^{m_d} \tilde{\bld a}_i\bld x+\tilde b_i.$$

(a)
 We first show that $C\subset \text{\textbf {dom}} (l(\bld a))$. Assume that $\bld a\in C$. That is $\bld a +\tilde {\bld a}_i\in \text{Newt} (p)$ for all $i=1,\dots,m_d$. Then, there are $\lambda_1,\dots,\lambda_{m_d}\geq 0$ with $\sum_{i=1}^{m_p}\lambda_1=1$ such that
 $$\bld a+\tilde {\bld a} _i= \sum_{i=1}^{n_p} \lambda_i  {\bld a}_i.$$
 Thus,
 $$(\bld a+\tilde {\bld a} _i)^T\bld x\leq \bigvee_{i=1}^{n_p} {\bld a}_i\bld x\leq p(\bld x)-\bigwedge_{i=1}^{n_p} b_i.$$
 Furthermore,
$$\tilde {\bld a}_i\bld x +\tilde b_i-\tilde b_i\geq \tilde {\bld a}_i\bld x +\tilde b_i-\bigvee_{j=1}^{m_d}\tilde b_j.$$
Combining the last two equations we have 
$$p(\bld x)-d(\bld x)-\bld a^T \bld x\geq\bigwedge_{i=1}^{n_p} b_i-\bigvee_{j=1}^{m_d}\tilde b_j>-\infty. $$
Hence, $\bld a \in \textbf{dom}(l)$.

Conversely assume that  $\bld a \in \textbf{dom}(l)$ but $\bld a \not\in C$. Then, there is a $\tilde {\bld a}_i$ such that $\bld a+\bld a_i\not \in \text{Newt}(p)$. Furthermore, since $\text{Newt}(p)$ is a convex set,  there is a vector $ \bld v$ such that
$(\bld a+\bld a_i)^T\bld v>\bld a_i^T\bld v$ for all $i=1,\dots,n_p$. Hence,
\begin{align*}
l(\bld a)&=\inf_{\bld x\in \mathbb R^n} \{p(\bld x)-d(\bld x)-\bld a^T \bld x\}\\
&\leq \lim_{ \lambda\rightarrow-\infty} p(\lambda\bld v)-d(\lambda\bld v)-\bld a^T( \lambda\bld v)=-\infty
\end{align*}
 Thus, $\bld a\not \in  \textbf{dom}(l)$, which is a contradiction. 

(b) Since 
$$q(\bld x)  = \sup_{\bld a \in \mathbb R^n} \{\bld a ^T\bld x-f^\star ({\bld a}) \}$$
It holds 
$$q(\bld x)=-\infty \Leftrightarrow l(\bld a)=-\infty, \text{ for all } \bld a   \Leftrightarrow C = \textbf{dom}(l)=\emptyset \hfill$$

(c) If  for some $i$ it holds  $\hat {\bld a}_i\not\in C$, then $l( \hat {\bld a}_i)=-\infty$. That is, for all $b$ there is an $\bld x\in\mathbb R^n$ with $\hat {\bld a}_i^T\bld x+b>f(\bld x)$. This is is particularly true for $b= \hat b_i$, and thus $q(\bld x)$ is not a quotient. \hfill

\subsection{Proof of Corollary \ref{Corollary_DimRed}} 
\label{Corollary_DimRedProof}

The proof of the first part is immediate from Proposition \ref{Prop_setC}. To contradict, assume that the inclusion in (ii) is not true. Then, there exists a vector $\bld x \in \text{span}\{\bld a_1,\dots,\bld a_{m_p}\}^\perp$ but $\bld x  \not\in\text{span}\{\hat{\bld a}_1,\dots,\hat{\bld a}_{m_q}\}^\perp$. Thus, there is an index $i_0$ such that $\hat{\bld a}_{i_0}^T\bld x\neq 0$. Without loss of generality assume that $\hat{\bld a}_{i_0}^T\bld x> 0$ (if it is negative, use $-\bld x$ in the place of $\bld x$). Using (i) we have $\lim_{\lambda\rightarrow\infty} (q(\lambda\bld x)+d(\lambda\bld x))=\infty$, but $\lim_{\lambda\rightarrow\infty} p(\lambda\bld x)<\infty$. Hence, there is a $\lambda $ such that $ p(\lambda\bld x)<q(\lambda\bld x)+d(\lambda\bld x)$, which contradicts the fact that $q$ is the quotient.

\subsection{Proof of Proposition \ref{Alg_prop}} 
\label{Proof4}
First we need to show that all the components of $\bld a^{E,z} $ are positive. 
 Due to the convexity of $E$, either $$\inf\{z:(\bld x,z)\in E\}=-\infty, \text{ for all } \bld x\in\mathbb R^n,$$  or
$$\inf\{z:(\bld x,z)\in E\}>-\infty, \text{ for all } \bld x\in\mathbb R^n.$$
If $E=\mathbb R^{n+1}$, then there are no non-trivial constraints, i.e., $L=0$. In the following assume that $L>0$, that is there are some nontrivial constraints, and $E\neq \mathbb R^{n+1}$.
From the definition of $E$, for all $\bld x\in\mathbb R^n$, it holds 
$\inf\{z:(\bld x,z)\in E\}<\infty,$ and $\sup\{z:(\bld x,z)\in E\}=\infty$.
 
 Note that it is not possible to have a constraint in \eqref{E_formula} in the form $[\bld A^{E,\bld x}]_l\bld x\geq \bld [b^E]_l $, that is nontrivial, i.e.,  $[\bld A^{E,\bld x}]_l\neq \bld 0^T$. Indeed, if such  a constraint existed, then for an $\bld x$ not satisfying this constraint we would have 
$\inf\{z:(\bld x,z)\in E\}=\infty$, which is a contradiction. On the other hand if there were a constraint with $[\bld A^{E,\bld x}]_l\bld x+[\bld a^{E,z}]_l z\geq [ \bld b^E]_l $, with $[\bld a^{E,z}]_l<0$ then, for all $(\bld x, z)\in E$ we would have $z\leq (-[\bld A^{E,\bld x}]_l\bld x\bld +[b^E]_l)/[\bld a^{E,z}]_l$	
and thus $\sup\{z:(\bld x,z)\in E\}<\infty$.

It is then easy to see that the function $q$ given by \eqref{q_formula} has as epigraph the set $E$. Therefore, it is the quotient.

To determine the remainder it is sufficient to find all the terms $i$ of $p(\bld x)$ for which $p(\bld x)>\tilde p(\bld x)$, for an $\bld x\in \mathbb R^n$  that the term $i$ attains the maximum in $p(\bld x)$. 
 The function $p(\bld x)-\tilde p(\bld x)$ is linear on $P_{i,j}$ for all $i,j$. Therefore, since $p(\bld x)-\tilde p(\bld x) \geq 0$, and $\bld x_{i,j}$ is in the relative interior of $P_{i,j}$ we have  either $p(\bld x)-\tilde p(\bld x)=0, \text{ for all }  \bld x\in P_{i,j}, \text{ or }  p(\bld x_{i,j})>\tilde p(\bld x_{i,j})$. Hence, the procedure described in Algorithm \ref{alg:ExactPolyDivi} produces the remainder.

\subsection{Proof of Proposition \ref{MonotProp}}
\label{Proof5}
 Observe that
$$p(\bld x_j) -d(\bld x_j)-q_t(\bld x_j)=f(\bld x_j) -q_t(\bld x_j)\geq 0,$$
where the last inequality is a consequence of the first constraint of \eqref{Prob2}. Furthermore,
\begin{align}
e(t) =\sum_{j=1}^{N} f(\bld x_j)- \sum_{j=1}^{N}\bigvee_{i=1}^{\tilde m_q} (\hat {\bld a}_{i,t}^T\bld x_j+\hat b_{i,t}),
\end{align}
where $\hat {\bld a}_{i,t}$ is the solution of \eqref{Prob3} at the iteration $t$.  
Each term of the last sum can be written as:
$$\bigvee_{i=1}^{\tilde m_q} (\hat {\bld a}_{i,t}^T\bld x_j+\hat b_{i,t}) =  (\hat {\bld a}_{i'_{j,t},t}^T\bld x_j+\hat b_{i'_{j,t},t}),$$
where $i'_{j,t}$ is the partition that $j$ belongs after step $t$, i.e., $j\in I_{i'_{j,t},t}$ and $I_{\cdot ,t}$ is the partition obtained at Step $6$ in iteration $t$.
Furthermore due to the optimization in \eqref{Prob2}, 
$$\sum_{j=1}^N\hat {\bld a}_{i'_{j,t},t}^T\bld x_j+\hat b_{i'_{j,t},t}\leq \sum_{j=1}^N\hat {\bld a}_{i'_{j,t},t+1}^T\bld x_j+\hat b_{i'_{j,t},t+1}.$$
From Step 6,
$$ \sum_{j=1}^N\hat {\bld a}_{i'_{j,t},t+1}^T\bld x_j+\hat b_{i'_{j,t},t+1}\leq \sum_{j=1}^N\hat {\bld a}_{i'_{j,t+1},t+1}^T\bld x_j+\hat b_{i'_{j,t+1},t+1}.$$
Therefore, 
$\sum_{j=1}^{N}\bigvee_{i=1}^{\tilde m_q} (\hat {\bld a}_{i,t}^T\bld x_j+\hat b_{i,t})$ is non-increasing, and thus, $e(t)$ is non-decreasing.

\mycomment{

\subsection{Proof of Proposition \ref{PropENFalg}}
\label{ProofPropENFalg}
For $p(\boldsymbol x) = \bigvee_{l=1}^{m_p} (\boldsymbol{a}_l^T\boldsymbol{x}+b_l) $,  function $\text{ENF}_{p}(\bld a)$  can be written as
$$\text{ENF}_{p}(\bld a) = \max_{\bld \lambda}\left\{   \sum_{l=1}^{m_p} \lambda_l b_l : \sum_{l=1}^{m_p} \lambda_l \bld a_l =\bld a              \right \}.$$
 Thus, since $\text{ENF}_{q+d}$ is a piecewise linear function, we have $\text{ENF}_{q+d}(\bld a)\leq \text{ENF}_{p}(\bld a)$,  for all $\bld a\in \mathbb R^n$, if and only if, for all terms $\hat {\bld a}^T\bld x+\hat b$ of $d(\bld x)$, there are constants $\lambda_{l,j}$ such that:
 $$\hat{\bld a}+\tilde{\bld a}_j = \sum_{l=1}^{m_p} \lambda_{l,j} \bld a_l, ~~\hat{b}+\tilde{b}_j \leq \sum_{l=1}^{m_p} \lambda_{l,j} b_l , ~~\sum_{l=1}^k \lambda_{j,l}=1.$$
}

%\mycomment
{

\subsection{Proof of Proposition \ref{IneqProp}}
\label{Proof6}
We first prove \eqref{Ineq1}. The quotients $Q(p_1,d)$ and $Q(p_2,d)$ satisfy
\begin{align}
p_1(\bld x)\geq d(\bld x)+Q(p_1,d)(\bld x),\label{ineq_p1}\\
p_2(\bld x)\geq d(\bld x)+Q(p_2,d)(\bld x)\label{ineq_p2}.
\end{align}
Thus,
\begin{align*}
p_1(\bld x)+p_2(\bld x)\geq d(\bld x)+[Q(p_1,d)+Q(p_2,d)(\bld x)+ d(\bld x)].
\end{align*}
But $Q(p_1+p_2,d)$ is the maximum tropical polynomial satisfying this inequality. Thus it is greater than or equal to the expression in the bracket, i.e., 
\begin{align*}
Q(p_1+p_2,d)(\bld x)\geq Q(p_1,d)(\bld x)+Q(p_2,d)(\bld x)+ d(\bld x).
\end{align*}

 To prove \eqref{Ineq2}, take the maximum of \eqref{ineq_p1} and  \eqref{ineq_p2}. Using the distributive property we get
\begin{align*}
p_1(\bld x)\vee p_2(\bld x)\geq d(\bld x)+[Q(p_1,d)(\bld x)\vee Q(p_2,d)(\bld x)].
\end{align*}
 Thus,
\begin{align*}
Q(p_1\vee p_2,d)\geq Q(p_1,d)(\bld x)\vee Q(p_2,d)(\bld x).
\end{align*}
  
We then prove \eqref{Ineq3}. The quotients $Q(p,d_1)$ and $Q(p,d_2)$ satisfy
\begin{align}
p(\bld x)\geq d_1(\bld x)+Q(p,d_1)(\bld x),\label{ineq_d1}\\
p(\bld x)\geq d_2(\bld x)+Q(p,d_2)(\bld x)\label{ineq_d2}.
\end{align}
 Thus,
\begin{align*}
p(\bld x)\geq d_1(\bld x)+d_2(\bld x)+[Q(p,d_1)(\bld x)+Q(p,d_2)(\bld x)-p(\bld x)].
\end{align*}
 Hence, 
\begin{align*}
Q(p,d_1+d_2)(\bld x)\geq Q(p,d_1)(\bld x)+ Q(p_2,d)(\bld x)-p(\bld x).
\end{align*}

 To prove \eqref{Ineq4}, take the maximum of \eqref{ineq_d1} and  \eqref{ineq_d2}
 \begin{align*}
p(\bld x)&\geq [d_1(\bld x)+Q(p,d_1)(\bld x)] \vee[d_2(\bld x)+Q(p,d_2)(\bld x)] \\&\geq  [d_1(\bld x)\vee d_2(\bld x)] + [Q(p,d_1)(\bld x)\wedge Q(p,d_2)(\bld x)]
\end{align*}
 Therefore, 
 $$Q(p,d_1\vee d_2)(\bld x)\geq Q(p,d_1)(\bld x)\wedge Q(p,d_2)(\bld x)$$
 }

 \subsection{Proof of Proposition \ref{Ineq_Comp_Quotient_Prop}}
\label{Comp_Comp_propos_proof}

The Newton polytopes of $p$ and $q$ are given by the zonotopes
\begin{align}
\text{Newt}(p) = \{\bld A \mu: \mu\in[0,1]^n \},\\
\text{Newt}(q) = \{\hat{\bld A} \lambda: \lambda\in[0,1]^{m_q} \},
\end{align}
where $\hat{\bld A}=[\hat{\bld a}_1~\dots~\hat{\bld a}_{m^q}].$ 
We start proving the following lemma.
\begin{lemma}
Let $\text{Newt}(p) $, $\text{Newt}(q) $ as the above. Then $\text{Newt}(q)\subset \text{Newt}(p) $ if and only if
\begin{align}
\bld 0 \leq \bld A^{-1} \hat{\bld a}_i,~~
\bld A^{-1} \sum_{i=1}^{m_q} \hat{\bld a}_i \leq \bld 1,
\label{lemmaNeqwtIneq}
\end{align}
 for $i=1,\dots,m_q$,
\end{lemma}
\textit{Proof of the lemma}:  $\text{Newt}(q)\subset \text{Newt}(p) $ if and only if for any $\bld \lambda\in[0,1]^{m_q}$ there exists a $\bld \mu\in[0,1]^n$ such that 
$\bld A\mu=\hat {\bld A} \bld \lambda.$
Thus, $\text{Newt}(q)\subset \text{Newt}(p) $  is equivalent to 
\begin{align}
\bld 0\leq \bld A^{-1} \hat {\bld A} \bld \lambda\leq \bld 1,
\label{Interm}
\end{align}
for all $\bld \lambda\in[0,1]^{m_q}$.

Direct part: Assume  $\text{Newt}(q)\subset \text{Newt}(p) $. Substituting  $\bld\lambda =\bld e_i$, i.e., the $i-$th unit vector, in \eqref{Interm}, we get the first inequality of \eqref{lemmaNeqwtIneq}. Now with $\bld\lambda =\bld1$ we get the second inequality of \eqref{lemmaNeqwtIneq}. 

Converse part: Assume that \eqref{lemmaNeqwtIneq} holds true. Multiplying the first inequality of \eqref{lemmaNeqwtIneq} by $\lambda_i\geq 0$ and adding, we get the first inequality of \eqref{Interm}. Furthermore, for  $0\leq \lambda_i\leq 1$, we have 
$\sum_{i=1}^{m_q} \lambda_i \bld A^{-1} \hat{\bld a}_i \leq\bld A^{-1} \sum_{i=1}^{m_q} \hat{\bld a}_i \leq \bld 1,$
which implies the right part of inequality of \eqref{Interm}. \hfill $\square$

From Lemma \ref{LemmaBackground},  $q(\bld x)\leq p(\bld x)$ for all $\bld x$ is equivalent to $\text{ENF}_q(\bld a)\leq \text{ENF}_p(\bld a)$, for all $\bld a$. 
We first compute $\text{ENF}_p(\bld a) $. It holds
\begin{align}
\label{ENFformula}
\text{ENF}_p(\bld a) = \bld B^T \bld A^{-1} \bld a,
\end{align}
for $\bld a\in \text{Newt}(p)$ and $-\infty $ otherwise. 
Indeed, the extended Newton function of $p$ is given by
 $\text{ENewt}_p (\bld a)= \max\{ \sum_{i=1}^n \mu_i b_i : \sum_{i=1}^n \mu_i\bld a_i=\bld a,\mu_i\in[0,1]\}.$
However, $\sum_{i=1}^n \mu_i\bld a_i=\bld a$ has a solution $\bld \mu =[\mu_1~\dots \mu_n]^T \in[0,1]^n$, if and only if $a\in \text{Newt}(p)$. Furthermore, for $a\in \text{Newt}(p)$ the solution is unique and given by $\bld \mu = \bld A^{-1}\bld a$. Thus $\text{ENF}_p(\bld a) $ is given by \eqref{ENFformula}.

The first two conditions of \eqref{PropIneqConditionsCompQuot} are equivalent to the inclusion 
$\text{Newt}(q)\subset \text{Newt}(p) $. 
To prove the direct part of the proposition observe that $\hat b_i\leq \text{ENF}_q(\hat{\bld a}_i) \leq \text{ENF}_p(\hat{\bld a}_i) = \bld B^T \bld A^{-1} \hat{\bld a}_i.$

To prove the converse part assume that \eqref{PropIneqConditionsCompQuot} holds true. Thus, since $\text{Newt}(q)\subset \text{Newt}(p)$, it holds $\text{ENF}_q(\hat{\bld a})\leq  \text{ENF}_p(\hat{\bld a})$, for all $\hat{\bld a}\not\in \text{Newt}(q)$. It remains to show that $\text{ENF}_q(\hat{\bld a})\leq \text{ENF}_p(\hat{\bld a})$, for all $\hat{\bld a}\in \text{Newt}(q)$. For such an $\hat{\bld a}$ it holds
\begin{align*}
\text{ENF}_q(\hat{\bld a}) &= \max\{\sum_{i=1}^{m_q} \lambda_i \hat b_i: \sum_{i=1}^{m_q}\lambda_i\hat {\bld a}_i =\hat{\bld a},\lambda_i\in[0,1]\}\\
&\leq \max\{\sum_{i=1}^{m_q} \lambda_i \bld B^T\bld A^{-1} \hat {\bld a}_i: \sum_{i=1}^{m_q}\lambda_i\hat {\bld a}_i =\hat{\bld a},\lambda_i\in[0,1]\}\\
& = \bld B^T\bld A^{-1} \hat {\bld a}   =\text{ENF}_p(\hat{\bld a}) 
\end{align*}

}
\bibliographystyle{IEEEtran}
\bibliography{refs}

\begin{IEEEbiography}[{\includegraphics[width=1in,height=1.25in,clip,keepaspectratio]{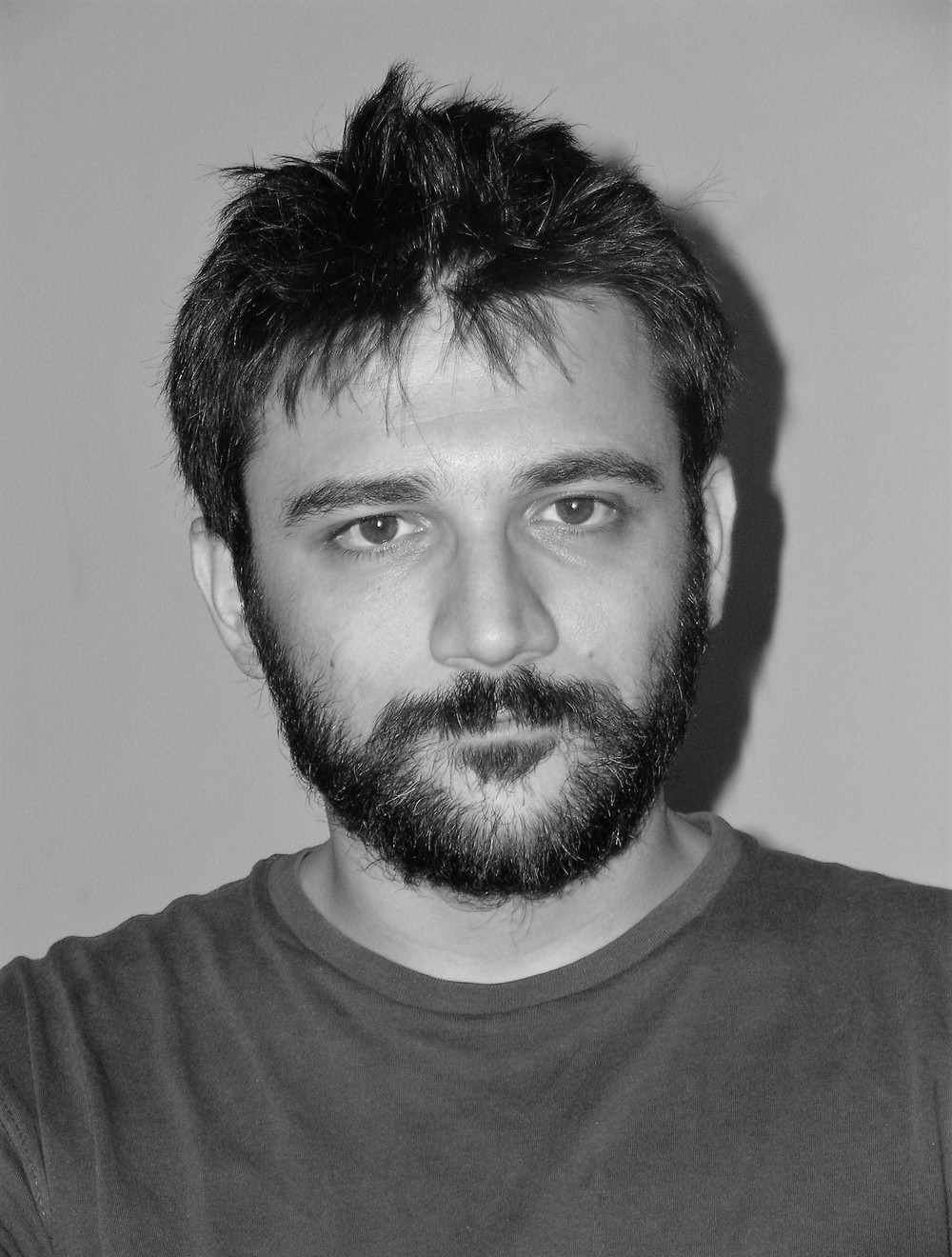}}]%
{Ioannis Kordonis}
 received both his  Diploma in Electrical and Computer Engineering and his Ph.D. degree from the National Technical
University of Athens, Greece, in 2009 and 2015 respectively. During 2016-2017 he was a post-doctoral researcher  in the University of Southern California and he is currently employed and during 2018-2019 he held  postdoctoral and ATER positions  at CentraleSup\'elec,  on the Rennes campus, in the Automatic Control Group - IETR. He is currently a postdoctoral researcher at the Intelligent Robotics and Automation Lab of the National Technical University of Athens. 

  His research interests include Game Theory, Stochastic Control theory, and Machine Learning. He is also interested in applications in the areas of Energy - Power Systems, Transportation Systems and in Bioengineering. 
\end{IEEEbiography}

\begin{IEEEbiography}[{\includegraphics[width=1in,height=1.25in,clip,keepaspectratio]{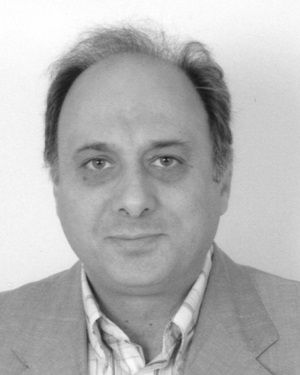}}]%
{Petros  Maragos}
 received the M.Eng. Diploma in E.E. from the National Technical University of Athens (NTUA) in 1980 and the M.Sc. and Ph.D. degrees from Georgia Tech, Atlanta, in 1982 and 1985. In 1985, he joined the faculty of the Division of Applied Sciences at Harvard University, where he worked for eight years as a professor of electrical engineering, affiliated with the Harvard Robotics Lab. In 1993, he joined the faculty of the School of ECE at Georgia Tech, affiliated with its Center for Signal and Image Processing. During periods of 1996-98, he had a joint appointment as director of research at the Institute of Language and Speech Processing in Athens. Since 1999, he has been working as professor at the NTUA School of ECE, where he is currently the director of the Intelligent Robotics and Automation Lab. He has held visiting positions at MIT in 2012 and at UPenn in 2016. He is a co-founder and since 2023 the acting director of the Institute of Robotics at the Athena Research Center. His research and teaching interests include signal processing, computer vision and speech, machine learning, and robotics. In the above areas as he has published numerous papers, book chapters, and co-edited three Springer research books. He has served as: Associate Editor for the IEEE Transactions on ASSP and the Transactions on PAMI, and as editorial board member for several journals on signal processing and computer vision; Co-organizer of several conferences, including recently ICASSP-2023 as general chair; Member of three IEEE SPS technical committees and the SPS Education Board; Member of the Greek National Council for Research and Technology. He is the recipient or co-recipient of several awards for his academic work, including a 1987-1992 US NSF Presidential Young Investigator Award;  1988 IEEE ASSP Young Author Best Paper Award; 1994 IEEE SPS Senior Best Paper Award;  1995 IEEE W.R.G. Baker Prize; 1996 Pattern Recognition Society's Honorable Mention best paper award; CVPR-2011 Gesture Recognition Workshop’s Best Paper Award; ECCVW-2020 Data Modeling Challenge Winner Award; CVPR-2022 best paper finalist. For his research contributions, he was elected Fellow of IEEE in 1995 and Fellow of EURASIP in 2010, and received the 2007 EURASIP Technical Achievement Award. He was elected IEEE SPS Distinguished Lecturer for 2017-2018. He is currently an IEEE Life Fellow. 

\end{IEEEbiography}

\end{document}